\pdfoutput=1 %
\documentclass[numbers]{xoople}

\usepackage[T1]{fontenc}
\usepackage{xspace}
\usepackage{booktabs}
\usepackage{array}  %
\usepackage{multirow}
\usepackage{subcaption}
\usepackage{xcolor}
\usepackage{graphicx}
\usepackage{amsmath,amssymb}

\newcommand{\paper}{\textsc{Analemma}\xspace}

\newcommand{\acq}[1]{\textbf{#1}}  %
\newcommand{\rel}[1]{\textbf{#1}}  %

\definecolor{passgreen}{HTML}{1F6F4A}
\definecolor{failred}{HTML}{B3261E}
\newcommand{\pass}[1]{{\color{passgreen}\textbf{#1}}}
\newcommand{\fail}[1]{{\color{failred}\textbf{#1}}}

\newcommand{\best}[1]{\textbf{#1}}
\newcommand{\second}[1]{\underline{#1}}

\newcommand{\tessera}{TESSERA\xspace}
\newcommand{\presto}{Presto\xspace}
\newcommand{\alphaearth}{AlphaEarth\xspace}
\newcommand{\dino}{DINOv3\xspace}
\newcommand{\mosaiks}{MOSAIKS\xspace}
\newcommand{\galileo}{Galileo\xspace}
\newcommand{\olmo}{OlmoEarth v1-Base\xspace}
\newcommand{\olmothree}{OlmoEarth v1.3-Base\xspace}
\newcommand{\prithvi}{Prithvi-EO 2.0\xspace}
\newcommand{\ramen}{RAMEN base\xspace}  %
\newcommand{\treesat}{TreeSatAI-TS\xspace}
\newcommand{\austria}{Austrian Crop\xspace}
\newcommand{\pastis}{PASTIS-R\xspace}
\newcommand{\ctm}{Crop Type Mapping\xspace}
\newcommand{\biomass}{BioMassters\xspace}
\newcommand{\borneo}{Borneo CHM\xspace}
\newcommand{\Stwo}{Sentinel-2\xspace}
\newcommand{\Sone}{Sentinel-1\xspace}

\renewcommand{\paragraph}[1]{\vspace{.4em}\noindent\textbf{#1.}}

\usepackage{placeins}
\usepackage{afterpage}
\providecommand\etal{\emph{et al}.\@\xspace}  %

\makeatletter
\newif\ifxp@narrow
\AtBeginEnvironment{figure}{\xp@narrowtrue}
\g@addto@macro\@floatboxreset{\ifxp@narrow\setlength{\linewidth}{0.6\textwidth}\centering\fi}
\newsavebox\xp@fitbox
\newif\ifxp@infit
\newif\ifxp@opened
\AtBeginEnvironment{table*}{%
  \let\xp@origtabular\tabular \let\xp@origendtabular\endtabular
  \renewenvironment{tabular}%
    {\ifxp@infit\xp@openedfalse\expandafter\xp@origtabular\else
       \xp@infittrue\xp@openedtrue\begin{lrbox}{\xp@fitbox}\expandafter\xp@origtabular\fi}%
    {\xp@origendtabular
     \ifxp@opened\end{lrbox}%
       \ifdim\wd\xp@fitbox>\linewidth\resizebox{\linewidth}{!}{\usebox\xp@fitbox}%
       \else\usebox\xp@fitbox\fi
     \fi}}
\makeatother

\title{\textsc{Analemma}\titlemark\hspace{0.4em}$\boldsymbol{\infty}$: A Temporal Benchmark for
  Geospatial Foundation Models}
\titlenote{The analemma is the figure-eight traced by the Sun in the sky when observed from
  one place at the same time of day over a year.}
\shorttitle{Analemma: A Temporal Benchmark for Geospatial Foundation Models}
\reporttype{Preprint}

\author{Alex L\'opez-Cifuentes}
\author{Laura Mart\'inez-Ferrer}
\author{Juan Ignacio Bravo P\'erez-Villar}

\date{November 25, 2026}
\correspondence{Alex L\'opez-Cifuentes, \email{alex.lopez@xoople.com}}
\metadata[data]{Data and code}{Data, harness and leaderboard will be publicly released soon}

\newcommand{\teaserblock}{%
  \begin{center}
    \vspace*{-6pt}
    \includegraphics[width=\textwidth]{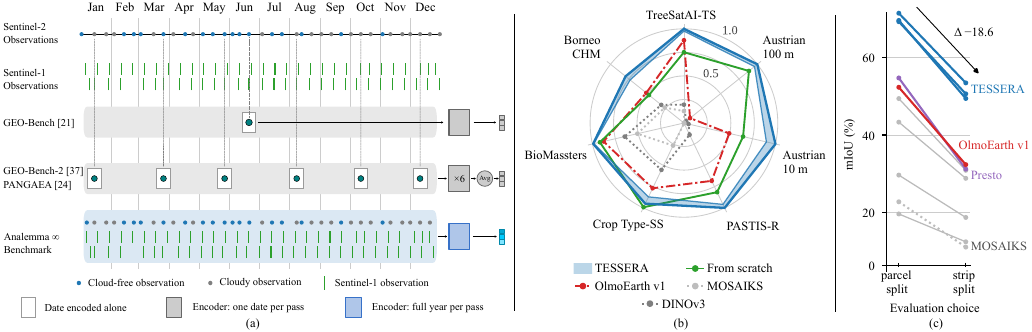}
    \captionof{figure}{(a)~Temporal input per pixel per year under each protocol: GEO-Bench supplies one date, GEO-Bench-2 and PANGAEA six dates encoded independently and averaged, \paper the full \Sone and \Stwo series with dates and cloud masks. (b)~Per-task normalized scores, one model per family. (c)~Effect of one unstated evaluation choice: mIoU on \austria 10~m segmentation at 30\,\% labels under two split rules. Scores drop by 7 to 23 points and three model pairs swap order. Further choices are quantified in Sec.~\ref{sec:repro}.}
    \label{fig:teaser}
    \vspace*{6pt}
  \end{center}
}

\begin{document}

\begin{abstract}
Satellite time series capture how land surfaces change over time. Pixel-wise temporal foundation models read a year of satellite observations
per pixel, yet public benchmarks lack a temporal axis or subsample
it to a few dates. We introduce \paper, a temporal benchmark of six datasets
that carry every 10 and 20~m \Stwo band and a time series spanning a growing season,
completed where their releases omit geometry, dates, or cloud masks.
One frozen-encoder protocol evaluates all models across label ratios and
three seeds, with two baselines without pretraining. The \tessera model ranks first overall. A from-scratch model on monthly
composites outranks every foundation model except \olmothree, suggesting that temporal input explains
most of the performance. Three evaluation choices left unstated by the source
publications each shift \tessera~v1 by 7.6 to 18.6 mIoU, against top-two gaps
of 0.6 to 2.8. Data, harness and leaderboard will be publicly released soon.
\end{abstract}
\maketitle

\afterpage{\begin{figure*}[!t]\teaserblock\end{figure*}}

\section{Introduction}
\label{sec:intro}

Sentinel-1 and Sentinel-2 image every point on land every few days, openly and at 10 and 20~m. Most variables that Earth observation (EO) monitors with these sensors, such as crop type or forest biomass, are defined by how they
change over months and require the full temporal record instead of a single acquisition. Geospatial foundation models, pretrained on large volumes of satellite imagery, have substantially improved performance on these tasks, and the methods proposed fall into three broad families. The first comprises single-image ViT-based encoders that produce patch-level embeddings from one or a few images~\cite{fuller2023croma,xiong2024dofa,reed2023scalemae,bastani2023satlas,hong2024spectralgpt}. The second comprises spatio-temporal cube encoders that attend jointly over space and time on a few composited frames~\cite{szwarcman2024prithvi,tseng2025galileo,herzog2025olmoearth}. The third family is more recent and consists of pixel-wise temporal encoders, which turn a year of observations from the \Stwo multispectral and \Sone radar satellites into one embedding per pixel~\cite{tseng2023presto,feng2025tessera,brown2025alphaearth,tseng2026olmoearthv13}. Their authors report recent gains on land use, land cover, and crop mapping~\cite{feng2025tessera,brown2025alphaearth,tseng2026olmoearthv13}. Many land cover classes are separable only by how they change during the growing season, so the number and spacing of dates a model reads sets a ceiling on what it can distinguish.

Several benchmarks have been proposed to evaluate geospatial foundation models, yet none scores a temporal encoder on a full year of input. GEO-Bench~\cite{lacoste2023geobench} has no temporal dimension: all twelve of its tasks are evaluated on a single date. GEO-Bench-2~\cite{simumba2025geobench2} and PANGAEA~\cite{marsocci2024pangaea} include multi-temporal datasets, but subsample the time axis to a handful of independently encoded frames (Fig.~\ref{fig:teaser}a). A model designed to read a full season or year is thus scored on a fraction of its input.
In the absence of a suitable suite, the papers introducing temporal encoders have evaluated them on ad hoc collections of their own~\cite{feng2025tessera,brown2025alphaearth,tseng2026olmoearthv13}. Many of these evaluations, however, leave open details needed to reproduce their results or to score a new model on the same data (Sec.~\ref{sec:repro}, Fig.~\ref{fig:teaser}c).

This paper addresses that gap. \paper comprises six datasets selected by two criteria: each must include every 10 and 20~m \Stwo band and a time series spanning at least one growing season. Where the original releases omit part of the temporal information, we complete them. Every model is evaluated under a single frozen-encoder protocol with a label-efficiency axis and three seeds per cell (one model trained on one dataset at one label ratio).

\paragraph{Contributions}
\begin{enumerate}
  \item \textbf{\paper, a temporal multi-band benchmark with a unified evaluation protocol,} including two baselines without pretraining, a random-feature floor and supervised models trained from scratch on composited raw bands, that show how much of each result is due to pretraining.
  \item \textbf{The first common evaluation of pixel-wise temporal encoders on a full year of input,} covering the three \tessera generations~\cite{feng2025tessera,feng2026tesserav2}, \presto~\cite{tseng2023presto}, \alphaearth~\cite{brown2025alphaearth}, and \olmothree~\cite{tseng2026olmoearthv13}, all under our protocol. It shows that \tessera leads (Fig.~\ref{fig:teaser}b), that a model trained from scratch on monthly composites ranks above every foundation model but one, and that pretraining adds its largest margin under label scarcity.
  \item \textbf{A systematic analysis of the sensitivity of published results to unstated evaluation choices,} and a public release of the completed data, the evaluation harness, and a leaderboard that accepts community-submitted embeddings with these choices fixed as code.
\end{enumerate}

\section{Related work}
\label{sec:related}

\paragraph{Geospatial foundation models} The three encoder families of Sec.~\ref{sec:intro} differ in how much spatial and temporal information they read. Beyond this families, recent work also extends the input along resolution and modality. AnySat~\cite{astruc2025anysat} accepts heterogeneous resolutions and modalities, Sentinel time series among them. Copernicus-FM~\cite{wang2025copernicusbench} and TerraMind~\cite{jakubik2025terramind} fuse several sensors per acquisition. Other single-image encoders target vision--language alignment~\cite{liu2024remoteclip} or multi-modal pretraining at scale~\cite{guo2024skysense}. Sec.~\ref{sec:protocol} gives the inclusion criterion of the benchmark.

\paragraph{Benchmarks for Earth observation} General-purpose suites rank encoders across many tasks under one protocol~\cite{lacoste2023geobench,simumba2025geobench2,marsocci2024pangaea,wang2025copernicusbench,fibaek2024phileo} and targeted benchmarks focus on one domain, such as forest monitoring~\cite{bountos2023fomobench} or sustainability~\cite{yeh2021sustainbench}, or one property, such as robustness~\cite{reobench} or spectral-band reliance~\cite{robinson2026similar}. These suites standardize the protocol per task, but the protocol itself limits the temporal input. Supp.~\ref{sup:suites} examines every dataset of the three suites closest to ours against the two entry criteria of \paper (Sec.~\ref{sec:benchmark}), which require every 10 and 20~m \Stwo band and a time series spanning at least one growing season, the input that multispectral temporal encoders are designed to read. None of the twelve GEO-Bench tasks and none of the nineteen GEO-Bench-2 datasets meet both, and three of the eleven PANGAEA datasets do. ChronoEarth benchmark~\cite{si2026chronoearth} is temporal but targets the EO-1 Hyperion hyperspectral archive, so its sensor and model sets do not overlap with ours.

\paragraph{Variance and reproducibility of evaluations} In machine learning, the variance due to data sampling, initialization and hyperparameter choice can be as large as the differences between the methods being ranked~\cite{bouthillier2021accounting,picard2021torch}, and reproducibility programs ask for code, splits and statistical spread to be reported~\cite{pineau2021improving}. In EO, recent audits examine whether released foundation models can be run and their metrics replicated~\cite{young2026usable}, how duplicated samples across splits of large geospatial datasets compromise evaluation~\cite{adimoolam2026leakage}, and how the composition of pretraining data changes downstream performance~\cite{kaur2026pretrain}. An audit of 152 geospatial foundation model papers finds 46 cross-paper disagreements of at least 10 points for the same model, benchmark and protocol, and calls for shared evaluations, baselines labeled as rerun or copied, variance reporting and a common harness~\cite{corley2026noone}. These works document the variance or locate it in training or in the data. We isolate a further source in Sec.~\ref{sec:repro} (protocol choices that neither the released code nor the paper determines) and compare the shift they cause with the gap between the models being ranked.

\begin{table*}[t]
  \centering
  \caption{Datasets in \paper, grouped by task. All provide \Sone and \Stwo
on a 10~m grid (20~m bands resampled) with acquisition dates and cloud masks, except \biomass, which is
supplied as monthly composites. License refers to the mirrored repository
(Sec.~\ref{sec:release}); origin denotes material whose source release
carries no license statement.}
  \label{tab:datasets}
  \scriptsize
  \setlength{\tabcolsep}{3pt}
  \renewcommand{\arraystretch}{1.12}
  \resizebox{\linewidth}{!}{%
  \begin{tabular}{@{}l l l l c l l c l l@{}}
    \toprule
    Task & Dataset & Domain & Region, year & Dates S2 / S1 & Sample & Train / Val / Test & Classes & Split rule & License \\
    \midrule
    \multirow{2}{*}{Classification}
      & \treesat~\cite{ahlswede2023treesatai} & Tree genera & Germany, 2017--20 & 90--146 / 239--323$^\P$ & 6$\times$6 px plot & 38\,520 / 6\,810 / 5\,044 & 15 & official & CC-BY-SA 4.0 \\
      & \austria @100 m~\cite{feng2025tessera} & Crop type & Austria, 2022 & 94 / 119 & pixel, 459$\times$518 scene & by parcel area & 17 & parcel & origin \\
    \midrule
    \multirow{2}{*}{Regression}
      & \biomass~\cite{nascetti2023biomassters} & Biomass & Finland, 2017--21 & 5--12 / 12 (monthly) & 256$^2$ px chip & 7\,820 / 869$^\dagger$ / 2\,773 & $\mathbb{R}_+$ (Mg\,ha$^{-1}$) & official & CC-BY 4.0 \\
      & \borneo~\cite{feng2025tessera,coomes2020danum} & Canopy height & Sabah, 2020 & 78 / 31$^\ddagger$ & 32$^2$ px patch & 300 / 100 / 100 & $\mathbb{R}_+$ (m) & by region & origin, CC-BY 4.0 \\
    \midrule
    \multirow{3}{*}{Segmentation}
      & \pastis~\cite{garnot2021pastis,garnot2022pastisr} & Crop type & France, 2018--19 & $\sim$43 / $\sim$135 & 128$^2$ px patch & 1\,455 / 482 / 496 & 19 & 5 official folds & CC-BY 4.0 \\
      & \austria @10 m & Crop type & Austria, 2022 & 144 / 119$^\S$ & 64$^2$ px patch & 3\,000 / 1\,000 / 1\,000 & 17 & strips & CC-BY 4.0 \\
      & \ctm~\cite{rustowicz2019croptype} & Crop type & South Sudan, 2017 & 49--216 / 25--147 & 64$^2$ px chip & 554 / 66 / 67 & 4 of 18 & official & CC-BY-SA 4.0 \\
    \bottomrule
  \end{tabular}}
  \\[2pt]\raggedright\scriptsize
  $^\dagger$ No published validation split; ours is a fixed 90/10 partition
of the training half under one seed, identical across all cells.
$^\ddagger$ \Sone descending orbit only.
$^\S$ Our 2022 series; the 94 dates at 100~m are the same solar days after
the reference's 90\,\% cloud filter (Fig.~\ref{fig:datasets}).
$^\P$ Series length varies by sample; values are medians per year, 2017
to 2020.
\end{table*}

\section{The \paper benchmark}
\label{sec:benchmark}

\paper only comprises datasets that carry every \Stwo band at 10 and 20~m
and a time series spanning at least one growing season, so that multispectral and temporal encoders receive the full input they were designed for.

\paragraph{Datasets}
Table~\ref{tab:datasets} summarizes the six selected datasets and Tab.~\ref{tab:releases} details what each release provides and what we completed (Supp.~\ref{sup:datasets}). Three datasets required only minor adjustments (\treesat, \ctm and \biomass). \treesat is used as released. For \ctm we use the original release rather than the PANGAEA repack, which drops the cloud masks and the \Sone orbit labels. For \biomass, we hold out a fixed 10\,\% of the training half as validation split, since the release publishes none. Neither \ctm nor \biomass provides sample locations.

The remaining three (\pastis, \austria, \borneo) required completion. We derived \pastis cloud masks from the \Stwo scene classification layer (SCL), a per-pixel map of cloud, shadow and surface classes. The \austria release contains a single scene at 100~m, with no geometry, dates, cloud masks or year. We located the scene, identified the year, acquired the 2022 \Sone and \Stwo series at 10~m, and rasterized 10~m labels from the open national parcel register. The \borneo release provides labels and the authors' embeddings but no imagery, georeference or dates. We recovered 482 of the 500 patch footprints and acquired the 2020 \Sone and \Stwo series over them.

\begin{table*}[t]
  \centering
  \caption{What each original release provides and what we completed. P: published with the original release. \acq{A}: acquired by us. P/\acq{A}: the release publishes one resolution and we acquired the other. \rel{R}: released as part of \paper. ---: unavailable. Two releases omit sample coordinates, which excludes coordinate-conditioned models from those datasets in Table~\ref{tab:main}.}
  \label{tab:releases}
  \small
  \setlength{\tabcolsep}{6pt}
  \begin{tabular}{@{}l cccccc cc@{}}
    \toprule
    & \multicolumn{6}{c}{Data} & \multicolumn{2}{c}{Protocol} \\
    \cmidrule(lr){2-7}\cmidrule(lr){8-9}
    Dataset & Imagery & Dates & Cloud masks & Geometry & Labels & Embeddings & Head & Results \\
    \midrule
    \treesat & P & P & P & P & P & \rel{R} & \rel{R} & \rel{R} \\
    \austria & P/\acq{A} & \acq{A}$^{c}$ & \acq{A} & \acq{A} & P/\acq{A} & P/\rel{R} & P/\rel{R} & P$^{a}$/\rel{R} \\
    \pastis & P & P & \acq{A} & P & P & \rel{R} & \rel{R} & \rel{R} \\
    \ctm & P & P & P & --- & P & \rel{R} & P/\rel{R} & P$^{b}$/\rel{R} \\
    \biomass & P & P & P & --- & P & P/\rel{R} & P/\rel{R} & P/\rel{R} \\
    \borneo & \acq{A} & \acq{A} & \acq{A} & \acq{A} & P & P/\rel{R} & P/\rel{R} & P$^{a}$/\rel{R} \\
    \bottomrule
  \end{tabular}
  \\[2pt]\raggedright\footnotesize $^{a}$\,Baseline embeddings only. $^{b}$\,Reading 6 frames of the series. $^{c}$\,Including the year of observation, which the release does not state.
\end{table*}

\paragraph{Data release and leaderboard}
\label{sec:release}
Each dataset is released as its own repository carrying the six items of Sec.~\ref{sec:repro}: per-sample footprints and acquisition dates, splits as index files, the training script with its head settings, the metric as code, the raw inputs under one band and unit contract, and every cell with its seed spread, together with the embeddings of every encoder. The leaderboard scores submitted embeddings with the same heads at a pinned protocol version. Layout, licenses and the submission flow are in Supp.~\ref{sup:release}.

\section{Evaluation protocol}
\label{sec:protocol}

\subsection{Setup}
\paragraph{Frozen encoders and precomputed embeddings}
We call every row of Tab.~\ref{tab:models} a model and each input variant (one date, annual or monthly composite) a configuration. Pretrained models run their encoder frozen, mapping each sample to embeddings on which only a light task-specific head is trained. The two baselines (MOSAIKS and From-scratch) have no encoder. Freezing is the operating regime of the pixel-wise temporal family and makes the label-efficiency sweep affordable. Each encoder runs once per checkpoint, input variant and dataset, and all results except those of \alphaearth come from our own encoding.

All GPU work ran on a single NVIDIA H100 NVL (94\,GiB), including the 1\,410 head fits behind the 470 cells reported. Per-pixel inference cost is compared in Fig.~\ref{fig:supp-cost}.

\paragraph{Downstream heads and label efficiency}
Classification tasks fit a per-pixel MLP on the embeddings (mean pooled on \treesat). Segmentation and regression tasks fit a U-Net decoder. Head architecture, split rule and evaluation metric are fixed per dataset and identical for every model. We select the best epoch on the validation split. To measure label efficiency, every model is evaluated on each task at different label ratios (Supp.~\ref{sup:protocol}). Classification is scored by weighted F1, segmentation by mean IoU and regression by RMSE. Each cell uses three seeds, and the tables report the mean and sample standard deviation.

\subsection{Evaluated models}
\label{sec:models}
\begin{table*}[t]
  \centering
  \caption{Models evaluated in the benchmark, grouped by input family. Temporal input describes what each model receives from our benchmark rather than what it can accept in principle. Output describes the embeddings extracted by each model. Parameters refer to the encoder as run frozen by the evaluation harness in every cell.}
  \label{tab:models}
  \small
  \setlength{\tabcolsep}{5pt}
  \begin{tabular}{@{}l l >{\raggedright\arraybackslash}m{4.0cm} l r@{}}
    \toprule
    Family & Model & Temporal input & Output & Params (M) \\
    \midrule
    \multirow{6}{*}{\begin{tabular}{@{}l@{}}Pixel-wise\\temporal\end{tabular}}
      & \tessera v1 / v1.1 / v2-M~\cite{feng2025tessera,feng2026tesserav2} & every acquisition, S1+S2 & pixel, 128/192/128-d & 87.8 / 57.7 / 21.0 \\
      & \presto~\cite{tseng2023presto} & 12 monthly composites + lat/lon & pixel, 128-d & 0.4 \\
      & \alphaearth~\cite{brown2025alphaearth} & annual product & pixel, 64-d & $\sim$480$^\ast$ \\
      & \olmothree~\cite{tseng2026olmoearthv13} & 12 monthly cloud-free mosaics, S1+S2 & pixel, 128-d & 128.3 \\
    \midrule
    \multirow{6}{*}{\begin{tabular}{@{}l@{}}Spatio-temporal\\cube\end{tabular}}
      & \galileo base~\cite{tseng2025galileo} & 12 monthly composites, S1+S2+NDVI + lat/lon & 4$\times$4 px token, 768-d & 86.5 \\
      & \olmo~\cite{herzog2025olmoearth} & 12 monthly composites, S1+S2 & 4$\times$4 px token, 1\,536-d & 88.9 \\
      & \prithvi 300M~\cite{szwarcman2024prithvi} & 4 quarterly composites, 6 optical bands & 16$\times$16 px token, 1\,024-d & 303.9 \\
      & \ramen~\cite{houdre2026ramen} & 12 monthly composites, S1+S2 & 4$\times$4 px token, 1\,536-d & 96.1 \\
    \midrule
    \multirow{3}{*}{Single image}
      & \dino (SAT)~\cite{simeoni2025dinov3} & one date; annual; monthly & 4$\times$4 px token, 4\,096-d & 6\,716 \\
      & \mosaiks~\cite{rolf2021mosaiks} & one date; annual; monthly & tile, 4\,096-d & 0 \\
      & From scratch & one date; annual; monthly & raw bands & 0 \\
    \bottomrule
  \end{tabular}
  \\[2pt]\raggedright\footnotesize $^\ast$ As reported by the authors~\cite{brown2025alphaearth}; the weights are not public, so it cannot be counted.
\end{table*}

\paragraph{Inclusion criterion}
The benchmark targets temporal encoders. A temporal encoder is included if (i) its weights or embeddings product has been released before the set of evaluated models was fixed in November 2026, (ii) it reads \Stwo as a series or as composites, and (iii) no sample of the six benchmark datasets, labeled or not, appears in its pretraining corpus, as documented by its authors, which excludes AnySat~\cite{astruc2025anysat}, whose corpus contains PASTIS-HD. Single-image encoders enter only as a reference, through one general encoder at scale and a random-feature floor (Tab.~\ref{tab:models}). \olmothree meets (iii) with one caveat: its authors select design choices by validation mIoU on PASTIS~\cite{tseng2026olmoearthv13}, whose \Stwo patches are those of our \pastis.

\paragraph{Pixel-wise temporal encoders}
\tessera is evaluated across three checkpoints (v1, v1.1 and v2), the architectural details are in Supp.~\ref{sup:protocol}. We produce \presto's monthly composites from our series, so its results are our reading of \presto, not a reproduction of the original pipeline. \alphaearth is a closed product, so we evaluate the annual product it releases, fetched at our sample locations, with our downstream heads, not an encoding we ran. It lacks full export coverage on \ctm and \biomass because neither dataset provides sample coordinates, and has no results on those datasets. \olmothree is run as its authors deploy it~\cite{tseng2026olmoearthv13}: 2$\times$2~px patches, one attention-pooler query per 10~m pixel, 16~px windows (160~m on the 10~m datasets), one cloud-free mosaic per month built from its least cloudy observations rather than a mean, and the distilled 128-d embedding.

\paragraph{Spatio-temporal cube encoders}
We set \galileo's climate, elevation and land-cover channels to absent.  \olmo and \olmothree receive only \Sone and the ten \Stwo bands at 10 and 20\,m. The rest of \Stwo bands, all Landsat bands and the map-product inputs in v1 are not included. \prithvi cannot be evaluated on \treesat due to its token size. \ramen uses the token grid of \olmo and receives only months with at least 95\,\% valid coverage.

\begin{table*}[t]
  \centering
  \caption{Main results on the \paper benchmark. Models are grouped by input family, and two label ratios are reported per task: the most restrictive (1\,\% of the training labels, or 10\,\% on \ctm) and the most permissive (all labels, or 30\,\% on \austria classification, whose split rule admits no higher ratio). Each cell reports the mean over three random seeds. Full results across all label ratios and standard deviations are in the supplementary material. Best per column in \best{bold}, second best \second{underlined}.}
  \label{tab:main}
  \scriptsize
  \setlength{\tabcolsep}{3.2pt}
  \resizebox{\linewidth}{!}{%
  \begin{tabular}{@{}ll cc cc cc cc cc cc cc@{}}
    \toprule
    & & \multicolumn{2}{c}{\treesat} & \multicolumn{2}{c}{\austria cls @100 m} & \multicolumn{2}{c}{\pastis} & \multicolumn{2}{c}{\austria seg @10 m} & \multicolumn{2}{c}{\ctm} & \multicolumn{2}{c}{\biomass} & \multicolumn{2}{c}{\borneo} \\
    & & \multicolumn{2}{c}{wF1 $\uparrow$} & \multicolumn{2}{c}{wF1 $\uparrow$} & \multicolumn{2}{c}{mIoU $\uparrow$} & \multicolumn{2}{c}{mIoU $\uparrow$} & \multicolumn{2}{c}{mIoU $\uparrow$} & \multicolumn{2}{c}{RMSE [Mg\,ha$^{-1}$] $\downarrow$} & \multicolumn{2}{c}{RMSE [m] $\downarrow$} \\
    \cmidrule(lr){3-4}\cmidrule(lr){5-6}\cmidrule(lr){7-8}\cmidrule(lr){9-10}\cmidrule(lr){11-12}\cmidrule(lr){13-14}\cmidrule(lr){15-16}
    Model & Input & 1\,\% & All & 1\,\% & 30\,\% & 1\,\% & All & 1\,\% & All & 10\,\% & All & 1\,\% & All & 1\,\% & All \\
    \midrule
    \tessera v1 & full year & \second{59.25} & \second{75.55} & \second{69.03} & 83.10 & 27.88 & \second{65.17} & 28.11 & \second{58.19} & 36.73$^{\dagger}$ & 59.72$^{\dagger}$ & 33.55 & \best{30.53} & 34.11 & \best{13.06} \\
    \tessera v1.1 & full year & 58.91 & 75.46 & 65.98 & \second{83.18} & \second{28.27} & 64.83 & \second{28.14} & 56.01 & 40.52$^{\dagger}$ & 59.69 & \second{33.32} & \second{30.60} & 34.31 & 13.25 \\
    \tessera v2-M & full year & \best{61.09} & \best{76.00} & \best{71.19} & \best{84.19} & \best{28.86} & \best{66.67} & \best{31.38} & \best{59.88}$^{\dagger}$ & 37.43$^{\dagger}$ & 60.85$^{\dagger}$ & 33.57 & 30.75 & 35.19 & \second{13.08} \\
    \alphaearth & annual product & 53.73 & 71.35 & 53.03 & 74.42 & 22.82$^{\dagger}$ & 54.96 & 15.33 & 35.44 & --- & --- & --- & --- & 34.27$^{\dagger}$ & 16.63 \\
    \presto & 12 months + lat/lon & 43.59 & 63.34 & 55.66 & 75.34 & 18.21$^{\dagger}$ & 51.42 & 14.50 & 37.18 & --- & --- & --- & --- & 35.18 & 15.01 \\
    \olmothree & 12 months & 49.12 & 69.44 & 38.62 & 61.29 & 24.73$^{\dagger}$ & 59.47 & 23.66 & 49.15 & \best{42.00}$^{\dagger}$ & 58.73$^{\dagger}$ & \best{32.74} & 30.86 & \second{31.44} & 13.76 \\
    \midrule
    \galileo base & 12 months + lat/lon & 41.59 & 62.46 & 23.98 & 32.43 & 16.21$^{\dagger}$ & 52.55 & 8.14 & 36.35 & --- & --- & --- & --- & 35.22 & 17.06 \\
    \olmo & 12 months & 52.38 & 71.36 & 23.20 & 31.40 & 19.28$^{\dagger}$ & 57.24 & 10.96 & 40.61 & 30.47$^{\dagger}$ & \second{60.86} & 37.76 & 32.94 & 34.31$^{\dagger}$ & 15.37 \\
    \prithvi 300M & 4 quarters & n/a & n/a & 23.50 & 26.40 & 12.29 & 31.28 & 6.39 & 16.19 & 18.15 & 43.80$^{\dagger}$ & 44.58 & 36.42 & 35.27 & 17.92 \\
    \ramen & 12 months & 40.24 & 59.68 & 26.01 & 32.03 & 20.00 & 52.78 & 12.17 & 37.09 & 25.02$^{\dagger}$ & 58.40$^{\dagger}$ & 43.39 & 39.39 & 35.04 & 18.64 \\
    \midrule
    \dino & one date & 20.02$^{\dagger}$ & 20.04 & 24.13 & 26.31 & 11.97 & 31.40 & 9.83 & 20.77 & 14.99 & 39.68$^{\dagger}$ & 55.93 & 48.55 & 34.68 & 18.02 \\
    \dino & best composite & 28.80 & 52.25 & 21.81 & 26.24 & 9.51$^{a}$ & 25.97$^{a}$ & 3.79$^{a}$ & 8.52$^{a}$ & 27.40 & 59.76 & 44.56$^{a}$ & 38.46$^{a}$ & 36.12 & 14.47 \\
    \mosaiks & one date & 15.26$^{\dagger}$ & 13.60 & 22.64 & 24.68 & 8.07 & 18.11 & 6.80 & 13.65 & 18.91 & 25.21 & 61.81 & 55.78 & \best{30.84}$^{\dagger}$ & 17.67 \\
    \mosaiks & best composite & 22.16 & 43.83 & 24.25 & 27.60 & 8.30$^{a}$ & 17.31$^{a}$ & 3.32$^{a}$ & 4.78$^{a}$ & 24.77$^{\dagger}$ & 49.73 & 47.64$^{a}$ & 42.81$^{a}$ & 35.54$^{a}$ & 17.04$^{a}$ \\
    \midrule
    From scratch & one date & 20.62$^{\dagger}$ & 37.10 & 40.53 & 48.15 & 9.05 & 39.04 & 10.47 & 37.50 & 19.85 & 28.29$^{\dagger}$ & 41.85 & 39.01 & 36.31 & 16.70 \\
    From scratch & monthly composite & 38.70 & 69.53 & 60.32 & 81.08 & 22.14 & 62.60 & 16.39 & 46.86 & \second{41.33}$^{\dagger}$ & \best{61.49}$^{\dagger}$ & 35.78 & 32.09 & 35.65 & 14.85 \\
    \bottomrule
  \end{tabular}
}
  \\[2pt]
  \raggedright\scriptsize $^a$ Annual composite. The monthly composite of \dino and \mosaiks (12\,$\times$\,4\,096 channels per pixel) was not computed on \pastis, \austria at 10~m and \biomass. On \borneo the annual composite is the better of the two for \mosaiks. $^\dagger$ Seed standard deviation above 2 points. ---: the model requires coordinates that the dataset does not provide. n/a: \prithvi's 16$\times$16~px token does not tile the 6$\times$6~px plots of \treesat.
\end{table*}

\paragraph{Single-image encoders and baselines}
\dino, in its satellite (SAT) variant, is pretrained on 493\,M RGB satellite images at 0.6\,m. \mosaiks samples random convolutional features from training folds only and defines the random-feature floor. The from-scratch model is the same head as the probe, an MLP or a U-Net, trained directly on the satellite bands; its variant on monthly composites adds both \Sone orbits. The annual composite is excluded from the crop datasets, where it averages out the growing season. The from-scratch model shares splits, seeds, label ratios and training budget with the frozen encoders, but trains with its own head settings (Supp.~\ref{sup:protocol}).

\section{\paper results}
\label{sec:results}

\begin{figure}[t]
  \centering
  \includegraphics[width=\linewidth]{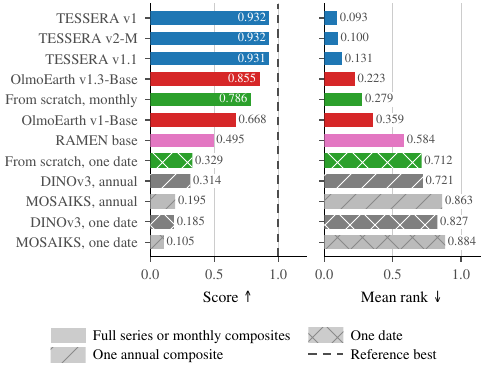}
  \caption{Global ranking on the \paper benchmark. Models evaluated across all six datasets, ordered by Score (frozen min-max normalization where 1 corresponds to the best reference) and by Mean Rank (where 0 denotes the top-ranked model). Models covering fewer datasets are compared in the supplementary material.}
  \label{fig:rank}
\end{figure}
\begin{figure*}[t]
  \centering
  \includegraphics[width=\textwidth]{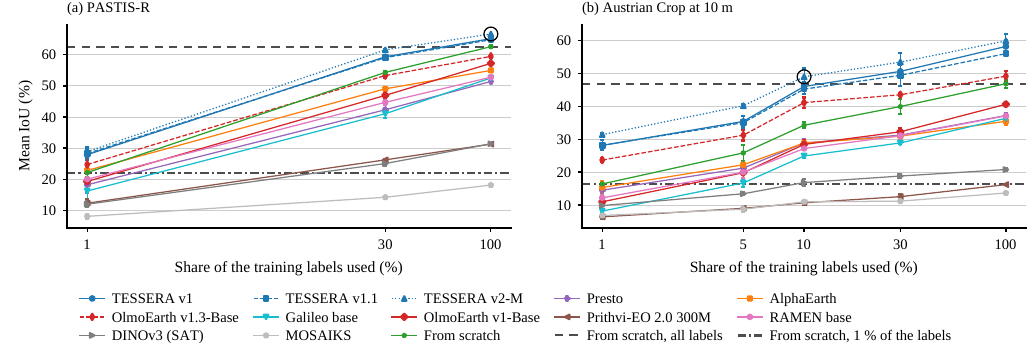}
  \caption{Label efficiency on (a) \pastis and (b) \austria at 10~m. Each point is the mean over three seeds, with error bars showing the standard deviation. The dashed line marks the best from-scratch model at full labels, and the dash-dotted line the same model at 1\,\% of the labels. The circle denotes the lowest ratio at which the best model surpasses the dashed reference.}
\label{fig:label-eff}
\end{figure*}
\subsection{Per-dataset results and global ranking}
Table~\ref{tab:main} reports the seven tasks at their most restrictive and most permissive label ratios, with the better of the annual and monthly composites for \dino and \mosaiks and the monthly composite for the from-scratch model (Sec.~\ref{sec:protocol}). Cells whose seed spread exceeds 2 points (five times the median spread of 0.4) are marked with $\dagger$; the complete sweep with standard deviations is in Supp.~\ref{sup:results}.

Comparing models across datasets requires aggregating results with different metrics, opposing signs, and varying label ratios. Following Bellemare \etal~\cite{bellemare2013arcade} and GEO-Bench~\cite{lacoste2023geobench}, each cell score $s_{m,(t,r)}$ is normalized to $[0, 1]$ by min-max scaling against frozen reference points (the lowest and highest values reached by any evaluated configuration in that cell, Supp.~\ref{sup:results}). %
The normalized scores are then averaged hierarchically so that each dataset weighs equally regardless of how many tasks or label ratios it contributes:
\begin{align}
  S_m &= \frac{1}{|\mathcal{D}|} \sum_{d} \frac{1}{|\mathcal{T}_d|} \sum_{t \in \mathcal{T}_d} \frac{1}{|\mathcal{R}_t|} \sum_{r \in \mathcal{R}_t} s_{m,(t,r)}. \label{eq:score}
\end{align}
Here $m$ is a model, $d$ a dataset with tasks $\mathcal{T}_d$, and $t$ a task with label ratios $\mathcal{R}_t$, so that a cell $(t,r)$ is one task at one label ratio. Reference points are frozen and released with the benchmark, so a future model can score above 1.

Fig.~\ref{fig:rank} reports the resulting global ranking. Only the twelve configurations evaluated on all six datasets are included. Models missing even one dataset would be averaged over a different subset, and a dataset is never missing at random. It is absent because the model lacks a required input, which correlates with performance. Fig.~\ref{fig:rank} also reports the mean rank, the position of each model within a cell scaled to 0 for the best and 1 for the worst, averaged as in Eq.~\eqref{eq:score}. It ignores how large a single cell's gap is and measures how consistently a model places near the top.

\subsection{Discussion of results}
\paragraph{A single model dominates the benchmark}
\tessera, the only encoder that receives every \Sone and \Stwo acquisition of the year (Tab.~\ref{tab:models}), achieves the best result in 10 of the 14 columns of Tab.~\ref{tab:main}, and its three generations occupy the top three positions in Fig.~\ref{fig:rank}. Two exceptions go to \olmothree, on \ctm at 10\,\% and on \biomass at 1\,\%. The from-scratch model leads \ctm at full labels, and \borneo at 1\,\% goes to \mosaiks, where the seed spread makes the comparison unreliable. %
\tessera v1 and v2-M score 0.932 against 0.855 for \olmothree, the next best pretrained encoder, with smaller encoders (87.8\,M and 21.0\,M vs.\ 128.3\,M parameters, Tab.~\ref{tab:models}) but 5.4 and 2.6 times its inference FLOPs (Fig.~\ref{fig:supp-cost}). The three \tessera generations are within 0.001 of each other in global score, so what separates them from the rest is what they share, not how they differ.

\presto, \alphaearth and \olmothree also receive the full year, so what sets \tessera apart is how it reads the year rather than access to it. It is the only evaluated encoder that combines three properties: (i) each 10\,m pixel is encoded from its own series alone, without mixing in its neighbors, (ii) the series is read acquisition by acquisition, with dates and cloud masks, rather than as monthly mosaics, and (iii) it carries every \Stwo and \Sone band. We hypothesize that this combination sets it apart, and no subset of them suffices: \presto has (i) and (iii) yet falls below the from-scratch model on most tasks, and \olmothree, which has (iii) and a per-pixel output but reads monthly mosaics through 2$\times$2~px patches, is the strongest encoder after \tessera. The clearest test of (i) is \austria at 100\,m, whose release keeps one pixel in every hundred, so that its samples lie 100\,m apart (Supp.~\ref{sup:datasets}). There, the encoders that mix neighboring pixels through patch tokens trail \tessera by 45 to 58 points, \olmothree by 23 to 33, and \presto by 9 to 16. These observations come from evaluation alone and do not isolate the three properties causally, but they are consistent with the design described by the \tessera authors~\cite{feng2025tessera}.

\begin{figure*}[t]
  \centering
  \includegraphics[width=\textwidth]{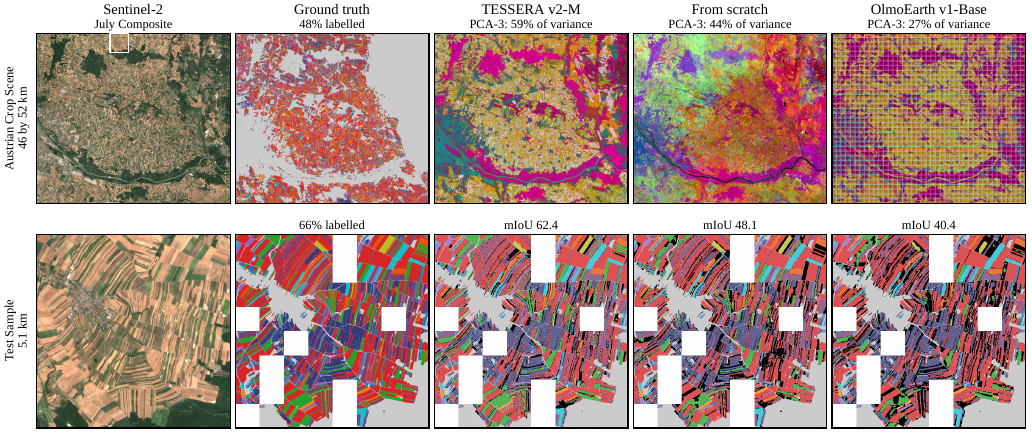}
  \caption{\textbf{Qualitative results on \austria at 10~m.} The task is crop-type segmentation: a good embedding separates neighboring parcels, and a good prediction follows parcel borders. The three models are the best row of each family at full labels in Tab.~\ref{tab:main}: \tessera v2-M, \olmo, and the from-scratch model on monthly composites. Top row: first three principal components of each embedding mapped to RGB over the full $46\times52$~km scene. The from-scratch panel shows principal components of its input composites (no embedding). \olmo is shown at its native $4\times4$~px token resolution (Supp.~\ref{sup:qual}). Colors are not comparable across panels. Bottom row: predictions on the outlined 5.1~km test sample with sample-level mIoU. Black: misclassified, gray: unlabeled, white: outside the test split.}
  \label{fig:qual}
\end{figure*}
\paragraph{Temporal input explains most of the performance}
Leading the benchmark appears to require these three properties together, but temporal input alone already accounts for most of the measured performance. A U-Net or MLP trained from-scratch on twelve monthly composites, with no pretrained encoder, lands within about four points of \tessera at full label availability on \austria classification, \pastis, \biomass, and \borneo, and ranks above every other pretrained encoder on these datasets, with three exceptions: \olmothree on \biomass and \borneo, and \dino on \borneo. It is also the cheapest configuration at its score (Fig.~\ref{fig:supp-cost}). The effect of temporal richness is visible even within this baseline: on \treesat the same MLP rises from 37.1 to 69.5 wF1 when switching from a single date to monthly composites, and a similar pattern holds on \pastis and \ctm. This dependence on temporal coverage is not specific to the baseline. Prior work shows that shortening the window \tessera reads from a year to a month costs 39\,\% of its segmentation accuracy on \pastis~\cite{anon2026temporal}.

\paragraph{Pretraining helps under label scarcity, mostly for \tessera and \olmothree}
At 1\,\% of the labels, \tessera leads the from-scratch model by 22.4 points on \treesat, 10.9 on \austria classification, and 6.7 on \pastis. The margin narrows at the most permissive ratio (6.5, 3.1, and 4.1, Tab.~\ref{tab:main}). Fig.~\ref{fig:label-eff} illustrates the same trend, on \austria at 10\,m, \tessera v2-M matches or exceeds the from-scratch model trained on all labels from 10\,\% of the labels onward, whereas on \pastis it does so only at full labels.

Only one other encoder shows this advantage. \olmothree leads the from-scratch model at 1\,\% on six of the seven tasks, by up to 10.4 points on \treesat, and its lead also narrows at the most permissive ratio, where it leads on three tasks by at most 2.3 points. Every other pretrained encoder falls below the from-scratch model at the most permissive ratio on at least four of the tasks it covers, including \dino, \prithvi, \olmo and \ramen. Scale and pretraining effort do not by themselves guarantee an advantage over a well-configured baseline, consistent with prior findings~\cite{marsocci2024pangaea}.

\subsection{Qualitative results}
In Fig.~\ref{fig:qual}, the \tessera embeddings resolve individual parcels, whereas the \olmo embeddings carry a 1.28~km grid inherited from their patch tokenization, and the from-scratch panel reflects only its input composites. %

\section{Sensitivity of published results to evaluation choices}
\label{sec:repro}

A published benchmark figure is the product of a model and a protocol. Where the protocol is fully stated, \paper reproduces published numbers closely (within 1.1 wF1 on \austria at 100~m for \tessera and \alphaearth, and within 1.1\,m RMSE on \borneo with all labels, Supp.~\ref{sup:validation}). Where part of the protocol is left unstated, any subsequent evaluation must infer it. Each of the three cases below shifts scores by several times the top-two gap, the distance between the two best models. Comparisons across publications therefore cannot establish a reliable ordering.

This is a structural property of how evaluations are published in EO, not of any single release~\cite{corley2026noone}. The papers we build on are among the most complete available, with open code, weights, and published results~\cite{feng2025tessera}, and 96.8\,\% of accessible geospatial foundation models release the code and data needed to replicate their metrics~\cite{young2026usable}. Yet in each case below, the published figures depend on choices that neither the code nor the paper determines. \paper fixes each of them as code (Tab.~\ref{tab:unstated}).

\textbf{Inconsistent split specification.} The \austria supplementary material~\cite{feng2025tessera} describes a parcel-based split for classification and a strip-based split for segmentation, while the released code implements only the former. Evaluating segmentation on the classification split yields figures up to 23.5 mIoU higher at 30\,\% of the labels (18.6 for \tessera v1), against a top-two gap of 2.8, with three model pairs swapping order (Fig.~\ref{fig:teaser}c). We adopt the strip rule, which reproduces the published figures~\cite{feng2025tessera} to within 0.25 points at 1\,\% of the labels for both \tessera and \presto.

\textbf{Ambiguous metric definition.} Mean IoU can be computed by pooling a single confusion matrix over the test split or by averaging per-sample scores. On \pastis~\cite{garnot2022pastisr}, the \tessera evaluation~\cite{feng2025tessera} does not specify which definition it employs, while the dataset's reference model, U-TAE~\cite{garnot2021pastis}, uses the pooled variant. From the same predictions with all labels, the two definitions diverge by 14.8 mIoU for \tessera v1 and by up to 15.6 across models, against a top-two gap of 1.5, so the published figures of \tessera and U-TAE compare distinct metrics and their ordering cannot be established.

\textbf{Unreported head configuration.} The head configuration for \pastis in the source evaluation~\cite{feng2025tessera} is not released. Two common settings, class weighting and embedding standardization, raise \tessera v2-M by 12.4 mIoU at 1\,\% of the labels (7.6 for \tessera v1) and lower it by 1.9 at full labels, with the change across models ranging from $-$0.2 to 12.4 against a top-two gap of 0.6. We report the default head, without either setting, because the published configuration is unknown. At 1\,\% of the labels this is the less favorable choice.

\textbf{Requirements for reproducible evaluation.} A benchmark on which a new model can be scored reproducibly must therefore ship the six items \paper releases (Sec.~\ref{sec:release}): a footprint and acquisition date per sample, the split as indices rather than prose, the complete training script of every head, the metric as code with its aggregation procedure, the raw inputs under a contract that fixes band order, units and resolution, and every result with its statistical spread and its experimental axes (label ratio, temporal extent, authorship of the run).

\section{Limitations and conclusions}
\label{sec:conclusion}

\subsection{Limitations}
All four limitations follow from the few public datasets that meet the entry criteria of \paper (Sec.~\ref{sec:benchmark}).

\textbf{Geographic coverage.} Four of the six datasets are European, with the remaining two in East Africa and Southeast Asia (Tab.~\ref{tab:datasets}). The Americas, temperate Asia, and the Southern Hemisphere are absent. Broader continental coverage is the primary axis for future expansion.

\textbf{Task diversity.} Four of the seven tasks involve crop type mapping, the domain where full-year sequences are most informative. The remaining tasks address tree genera classification and forest structure regression. Phenology-driven tasks favour temporal encoders, but targets with weaker seasonality (urban land use, water body delineation, disturbance mapping) are not yet represented. All six datasets are Sentinel-only, so encoders built for Landsat, very-high-resolution or hyperspectral imagery are read outside their native input.

\textbf{Dataset provenance.} Five of the six datasets were among the evaluation tasks of \tessera~\cite{feng2025tessera}, and two were released with it. The other three are established temporal datasets curated by independent groups. On \ctm, the one dataset outside that evaluation, \tessera does not lead: \olmothree ranks first at 10\,\% and the from-scratch model with all labels. On the four datasets not released with \tessera, its three generations still rank first by Score (0.962 to 0.970, against 0.902 for \olmothree). Adding datasets from sources unrelated to any evaluated model would further test the generality of the rankings.

\textbf{Measurement precision.} On \ctm the standard deviation over seeds reaches approximately 10 mIoU at full labels, and on \borneo at 1\,\% only three training patches remain with a spread of $\pm$8.6 RMSE. Three of the four exceptions to the lead of \tessera (Sec.~\ref{sec:results}) fall in these two datasets. The fourth, \biomass at 1\,\%, goes to \olmothree by 0.6 RMSE with a seed spread of 0.1. Because every seed is released, a later model can be compared within that spread rather than against a single run.

\subsection{Conclusions}
\paper is the first benchmark that scores temporal encoders on a full year of \Sone and \Stwo observations, under one frozen-encoder protocol with a label-efficiency axis and two baselines. On it, \tessera ranks first, \olmothree second, and a model trained from scratch on monthly composites ranks above every remaining foundation model. Temporal input thus explains most of the measured performance, and pretraining adds its largest margin under label scarcity.

The three unstated protocol choices we identify each shift published figures by several times the gap between the top-ranked models. For this reason, we release the benchmark whole, with its protocol as code, rather than as a table of numbers. A new temporal encoder can now be scored on a full year of input, against every encoder of this study and two baselines, with each choice that moved published figures fixed in advance. Data, harness, reference points and leaderboard (Sec.~\ref{sec:release}) will be made publicly available soon.

\bibliographystyle{ieeenat_fullname}
\bibliography{main}

\begin{thebibliography}{46}
\providecommand{\natexlab}[1]{#1}
\providecommand{\url}[1]{\texttt{#1}}
\expandafter\ifx\csname urlstyle\endcsname\relax
  \providecommand{\doi}[1]{doi: #1}\else
  \providecommand{\doi}{doi: \begingroup \urlstyle{rm}\Url}\fi

\bibitem[Adimoolam et~al.(2026)Adimoolam, Poullis, and
  Averkiou]{adimoolam2026leakage}
Yeshwanth~Kumar Adimoolam, Charalambos Poullis, and Melinos Averkiou.
\newblock Data leakage detection and de-duplication in large scale geospatial
  image datasets.
\newblock In \emph{CVPR}, pages 72--81, 2026.

\bibitem[Ahlswede et~al.(2023)Ahlswede, Schulz, Gava, Helber, Bischke,
  F{\"o}rster, Arias, Hees, Demir, and Kleinschmit]{ahlswede2023treesatai}
Steve Ahlswede, Christian Schulz, Christiano Gava, Patrick Helber, Benjamin
  Bischke, Michael F{\"o}rster, Florencia Arias, J{\"o}rn Hees, Beg{\"u}m
  Demir, and Birgit Kleinschmit.
\newblock {TreeSatAI} benchmark archive: a multi-sensor, multi-label dataset
  for tree species classification in remote sensing.
\newblock \emph{Earth System Science Data}, 15\penalty0 (2):\penalty0 681--695,
  2023.

\bibitem[Astruc et~al.(2025)Astruc, Gonthier, Mallet, and
  Landrieu]{astruc2025anysat}
Guillaume Astruc, Nicolas Gonthier, Cl{\'e}ment Mallet, and Loic Landrieu.
\newblock {AnySat}: One {Earth} observation model for many resolutions, scales,
  and modalities.
\newblock In \emph{CVPR}, pages 19530--19540, 2025.

\bibitem[Bastani et~al.(2023)Bastani, Wolters, Gupta, Ferdinando, and
  Kembhavi]{bastani2023satlas}
Favyen Bastani, Piper Wolters, Ritwik Gupta, Joe Ferdinando, and Aniruddha
  Kembhavi.
\newblock {SatlasPretrain}: A large-scale dataset for remote sensing image
  understanding.
\newblock In \emph{ICCV}, pages 16726--16736. IEEE, 2023.

\bibitem[Bellemare et~al.(2013)Bellemare, Naddaf, Veness, and
  Bowling]{bellemare2013arcade}
Marc~G. Bellemare, Yavar Naddaf, Joel Veness, and Michael Bowling.
\newblock The arcade learning environment: An evaluation platform for general
  agents.
\newblock \emph{Journal of Artificial Intelligence Research}, 47:\penalty0
  253--279, 2013.

\bibitem[Bountos et~al.(2025)Bountos, Ouaknine, Papoutsis, and
  Rolnick]{bountos2023fomobench}
Nikolaos~Ioannis Bountos, Arthur Ouaknine, Ioannis Papoutsis, and David
  Rolnick.
\newblock {FoMo}: Multi-modal, multi-scale and multi-task remote sensing
  foundation models for forest monitoring.
\newblock In \emph{AAAI}, pages 27858--27868. Association for the Advancement
  of Artificial Intelligence (AAAI), 2025.

\bibitem[Bouthillier et~al.(2021)Bouthillier, Delaunay, Bronzi, Trofimov,
  Nichyporuk, Szeto, Mohammadi~Sepahvand, Raff, Madan, Voleti, Ebrahimi~Kahou,
  Michalski, Arbel, Pal, Varoquaux, and Vincent]{bouthillier2021accounting}
Xavier Bouthillier, Pierre Delaunay, Mirko Bronzi, Assya Trofimov, Brennan
  Nichyporuk, Justin Szeto, Nazanin Mohammadi~Sepahvand, Edward Raff, Kanika
  Madan, Vikram Voleti, Samira Ebrahimi~Kahou, Vincent Michalski, Tal Arbel,
  Chris Pal, Gael Varoquaux, and Pascal Vincent.
\newblock Accounting for variance in machine learning benchmarks.
\newblock In \emph{Proceedings of Machine Learning and Systems}, pages
  747--769, 2021.

\bibitem[Brown et~al.(2025)Brown, Kazmierski, Pasquarella, Rucklidge,
  Samsikova, Zhang, Shelhamer, Lahera, Wiles, Ilyushchenko, Gorelick, Zhang,
  Alj, Schechter, Askay, Guinan, Moore, Boukouvalas, and
  Kohli]{brown2025alphaearth}
Christopher~F. Brown, Michal~R. Kazmierski, Valerie~J. Pasquarella, William~J.
  Rucklidge, Masha Samsikova, Chenhui Zhang, Evan Shelhamer, Estefania Lahera,
  Olivia Wiles, Simon Ilyushchenko, Noel Gorelick, Lihui~Lydia Zhang, Sophia
  Alj, Emily Schechter, Sean Askay, Oliver Guinan, Rebecca Moore, Alexis
  Boukouvalas, and Pushmeet Kohli.
\newblock {AlphaEarth} foundations: An embedding field model for accurate and
  efficient global mapping from sparse label data, 2025.

\bibitem[Coomes and Jackson(2022)]{coomes2020danum}
David Coomes and Toby Jackson.
\newblock Airborne {LiDAR} and {RGB} imagery from {Sepilok Reserve} and {Danum
  Valley} in {Malaysia} in 2020.
\newblock Dataset, 2022.

\bibitem[Corley et~al.(2026)Corley, Lehmann, Robinson, Tseng, Fuller,
  Alemohammad, Shelhamer, Marcus, and Kerner]{corley2026noone}
Isaac Corley, Nils Lehmann, Caleb Robinson, Gabriel Tseng, Anthony Fuller,
  Hamed Alemohammad, Evan Shelhamer, Jennifer Marcus, and Hannah Kerner.
\newblock No one knows the state of the art in geospatial foundation models.
\newblock In \emph{NeurIPS}, 2026.

\bibitem[Feng et~al.(2026{\natexlab{a}})Feng, Atzberger, Jaffer, Knezevic,
  Sormunen, Young, Lisaius, Immitzer, Jackson, Ball, Coomes, Madhavapeddy,
  Blake, and Keshav]{feng2025tessera}
Zhengpeng Feng, Clement Atzberger, Sadiq Jaffer, Jovana Knezevic, Silja
  Sormunen, Robin Young, Madeline~C. Lisaius, Markus Immitzer, Toby Jackson,
  James Ball, David~A. Coomes, Anil Madhavapeddy, Andrew Blake, and Srinivasan
  Keshav.
\newblock {TESSERA}: Temporal embeddings of surface spectra for earth
  representation and analysis.
\newblock In \emph{CVPR}, pages 34818--34831, 2026{\natexlab{a}}.

\bibitem[Feng et~al.(2026{\natexlab{b}})Feng, Jaffer, Shokar, Knezevic, Ball,
  Sousa, Elvers, Lisaius, Atzberger, Young, Naik, Robinson, Coomes,
  Madhavapeddy, and Keshav]{feng2026tesserav2}
Zhengpeng Feng, Sadiq Jaffer, Ira Shokar, Jovana Knezevic, James Ball, Pedro
  Sousa, Mark Elvers, Madeline Lisaius, Clement Atzberger, Robin Young, Aneesh
  Naik, Niall Robinson, David Coomes, Anil Madhavapeddy, and Srinivasan Keshav.
\newblock {TESSERA} v2: Scaling pixel-wise earth foundation models,
  2026{\natexlab{b}}.

\bibitem[Fibaek et~al.(2024)Fibaek, Camilleri, Luyts, Dionelis, and
  Le~Saux]{fibaek2024phileo}
Casper Fibaek, Luke Camilleri, Andreas Luyts, Nikolaos Dionelis, and Bertrand
  Le~Saux.
\newblock {PhilEO} bench: Evaluating geo-spatial foundation models.
\newblock In \emph{{IGARSS} 2024 -- 2024 {IEEE} International Geoscience and
  Remote Sensing Symposium}, pages 2739--2744. IEEE, 2024.

\bibitem[Fuller et~al.(2023)Fuller, Millard, and Green]{fuller2023croma}
Anthony Fuller, Koreen Millard, and James~R. Green.
\newblock {CROMA}: Remote sensing representations with contrastive
  radar-optical masked autoencoders.
\newblock In \emph{NeurIPS}, pages 5506--5538. Neural Information Processing
  Systems Foundation, Inc. (NeurIPS), 2023.

\bibitem[Guerrero-Viu et~al.(2026)Guerrero-Viu, L{\'o}pez-Cifuentes,
  P{\'e}rez-Villar, and Pacifici]{anon2026temporal}
Julia Guerrero-Viu, Alex L{\'o}pez-Cifuentes, Ignacio P{\'e}rez-Villar, and
  Fabio Pacifici.
\newblock Temporal sensitivity analysis of {Tessera} embeddings, 2026.

\bibitem[Guo et~al.(2024)Guo, Lao, Dang, Zhang, Yu, Ru, Zhong, Huang, Wu, Hu,
  He, Wang, Chen, Yang, Zhang, and Li]{guo2024skysense}
Xin Guo, Jiangwei Lao, Bo Dang, Yingying Zhang, Lei Yu, Lixiang Ru, Liheng
  Zhong, Ziyuan Huang, Kang Wu, Dingxiang Hu, Huimei He, Jian Wang, Jingdong
  Chen, Ming Yang, Yongjun Zhang, and Yansheng Li.
\newblock {SkySense}: A multi-modal remote sensing foundation model towards
  universal interpretation for earth observation imagery.
\newblock In \emph{CVPR}, pages 27662--27673. IEEE, 2024.

\bibitem[Herzog et~al.(2026)Herzog, Bastani, Zhang, Tseng, Redmon, Sablon,
  Park, Morrison, Buraczynski, Farley, Hansen, Howe, Johnson, Otterlee,
  Schmitt, Pitelka, Daspit, Ratner, Wilhelm, Wood, Jacobi, Kerner, Shelhamer,
  Farhadi, Krishna, and Beukema]{herzog2025olmoearth}
Henry Herzog, Favyen Bastani, Yawen Zhang, Gabriel Tseng, Joseph Redmon,
  Hadrien Sablon, Ryan Park, Jacob Morrison, Alexandra Buraczynski, Karen
  Farley, Josh Hansen, Andrew Howe, Patrick~Alan Johnson, Mark Otterlee, Ted
  Schmitt, Hunter Pitelka, Stephen Daspit, Rachel Ratner, Christopher Wilhelm,
  Sebastian Wood, Mike Jacobi, Hannah Kerner, Evan Shelhamer, Ali Farhadi,
  Ranjay Krishna, and Patrick Beukema.
\newblock {OlmoEarth}: Stable latent image modeling for multimodal {Earth}
  observation.
\newblock In \emph{CVPR}, pages 34806--34817, 2026.

\bibitem[Hong et~al.(2024)Hong, Zhang, Li, Li, Li, Yao, Yokoya, Li, Ghamisi,
  Jia, Plaza, Gamba, Benediktsson, and Chanussot]{hong2024spectralgpt}
Danfeng Hong, Bing Zhang, Xuyang Li, Yuxuan Li, Chenyu Li, Jing Yao, Naoto
  Yokoya, Hao Li, Pedram Ghamisi, Xiuping Jia, Antonio Plaza, Paolo Gamba,
  Jon~Atli Benediktsson, and Jocelyn Chanussot.
\newblock {SpectralGPT}: Spectral remote sensing foundation model.
\newblock \emph{IEEE TPAMI}, 46\penalty0 (8):\penalty0 5227--5244, 2024.

\bibitem[Houdr{\'e} et~al.(2026)Houdr{\'e}, Marcos, Riffaud~de Turckheim,
  Ienco, Wendling, Kurtz, and Lobry]{houdre2026ramen}
Nicolas Houdr{\'e}, Diego Marcos, Hugo Riffaud~de Turckheim, Dino Ienco,
  Laurent Wendling, Camille Kurtz, and Sylvain Lobry.
\newblock {RAMEN}: Resolution-adjustable multimodal encoder for {Earth}
  observation.
\newblock In \emph{CVPR}, pages 27838--27848, 2026.

\bibitem[Jakubik et~al.(2025)Jakubik, Yang, Blumenstiel, Scheurer, Sedona,
  Maurogiovanni, Bosmans, Dionelis, Marsocci, Kopp, Ramachandran, Fraccaro,
  Brunschwiler, Cavallaro, Bernabe-Moreno, and
  Long{\'e}p{\'e}]{jakubik2025terramind}
Johannes Jakubik, Felix Yang, Benedikt Blumenstiel, Erik Scheurer, Rocco
  Sedona, Stefano Maurogiovanni, Jente Bosmans, Nikolaos Dionelis, Valerio
  Marsocci, Niklas Kopp, Rahul Ramachandran, Paolo Fraccaro, Thomas
  Brunschwiler, Gabriele Cavallaro, Juan Bernabe-Moreno, and Nicolas
  Long{\'e}p{\'e}.
\newblock {TerraMind}: Large-scale generative multimodality for {Earth}
  observation.
\newblock In \emph{ICCV}, pages 7383--7394, 2025.

\bibitem[Kaur et~al.(2026)Kaur, Purohit, Muhawenayo, Rolf, and
  Kerner]{kaur2026pretrain}
Amandeep Kaur, Mirali Purohit, Gedeon Muhawenayo, Esther Rolf, and Hannah
  Kerner.
\newblock Pretrain where? {I}nvestigating how pretraining data diversity
  impacts geospatial foundation model performance.
\newblock In \emph{CVPRW}, pages 7949--7959, 2026.

\bibitem[Lacoste et~al.(2023)Lacoste, Lehmann, Rodriguez, Sherwin, Kerner,
  L{\"u}tjens, Irvin, Dao, Alemohammad, Drouin, Gunturkun, Huang, Vazquez,
  Newman, Bengio, Ermon, and Zhu]{lacoste2023geobench}
Alexandre Lacoste, Nils Lehmann, Pau Rodriguez, Evan~David Sherwin, Hannah
  Kerner, Bj{\"o}rn L{\"u}tjens, Jeremy~Andrew Irvin, David Dao, Hamed
  Alemohammad, Alexandre Drouin, Mehmet Gunturkun, Gabriel Huang, David
  Vazquez, Dava Newman, Yoshua Bengio, Stefano Ermon, and Xiao~Xiang Zhu.
\newblock {GEO-Bench}: Toward foundation models for earth monitoring.
\newblock In \emph{NeurIPS}, pages 51080--51093. Neural Information Processing
  Systems Foundation, Inc. (NeurIPS), 2023.

\bibitem[Li et~al.(2025)Li, Tao, Zhang, Liu, Xiong, Luo, Liu, Pechenizkiy, Zhu,
  and Huang]{reobench}
Xiang Li, Yong Tao, Siyuan Zhang, Siwei Liu, Zhitong Xiong, Chunbo Luo, Lu Liu,
  Mykola Pechenizkiy, Xiao~Xiang Zhu, and Tianjin Huang.
\newblock {REOBench}: Benchmarking robustness of earth observation foundation
  models.
\newblock In \emph{NeurIPS}, pages 68313--68334. Neural Information Processing
  Systems Foundation, Inc. (NeurIPS), 2025.

\bibitem[Liu et~al.(2024)Liu, Chen, Guan, Zhou, Zhu, Ye, Fu, and
  Zhou]{liu2024remoteclip}
Fan Liu, Delong Chen, Zhangqingyun Guan, Xiaocong Zhou, Jiale Zhu, Qiaolin Ye,
  Liyong Fu, and Jun Zhou.
\newblock {RemoteCLIP}: A vision language foundation model for remote sensing.
\newblock \emph{{IEEE} Transactions on Geoscience and Remote Sensing},
  62:\penalty0 1--16, 2024.

\bibitem[Marsocci et~al.(2026)Marsocci, Jia, Le~Bellier, Kerekes, Zeng, Hafner,
  Gerard, Brune, Yadav, Shibli, Fang, Ban, Vergauwen, Audebert, and
  Nascetti]{marsocci2024pangaea}
Valerio Marsocci, Yuru Jia, Georges Le~Bellier, David Kerekes, Liang Zeng,
  Sebastian Hafner, Sebastian Gerard, Eric Brune, Ritu Yadav, Ali Shibli, Heng
  Fang, Yifang Ban, Maarten Vergauwen, Nicolas Audebert, and Andrea Nascetti.
\newblock {PANGAEA}: Assessing geospatial foundation models capabilities
  through a global and inclusive benchmark.
\newblock \emph{{IEEE} Geoscience and Remote Sensing Magazine}, 14\penalty0
  (1):\penalty0 245--285, 2026.

\bibitem[McInnes et~al.(2018)McInnes, Healy, and Melville]{mcinnes2018umap}
Leland McInnes, John Healy, and James Melville.
\newblock {UMAP}: Uniform manifold approximation and projection for dimension
  reduction, 2018.

\bibitem[Nascetti et~al.(2023)Nascetti, Yadav, Brodt, Qu, Fan, Shendryk, Shah,
  and Chung]{nascetti2023biomassters}
Andrea Nascetti, Ritu Yadav, Kirill Brodt, Qixun Qu, Hongwei Fan, Yuri
  Shendryk, Isha Shah, and Christine Chung.
\newblock {BioMassters}: A benchmark dataset for forest biomass estimation
  using multi-modal satellite time-series.
\newblock In \emph{NeurIPS}, pages 20409--20420. Neural Information Processing
  Systems Foundation, Inc. (NeurIPS), 2023.

\bibitem[Picard(2021)]{picard2021torch}
David Picard.
\newblock \texttt{torch.manual\_seed(3407)} is all you need: On the influence
  of random seeds in deep learning architectures for computer vision, 2021.

\bibitem[Pineau et~al.(2021)Pineau, Vincent-Lamarre, Sinha, Larivi{\`e}re,
  Beygelzimer, d'Alch{\'e} Buc, Fox, and Larochelle]{pineau2021improving}
Joelle Pineau, Philippe Vincent-Lamarre, Koustuv Sinha, Vincent Larivi{\`e}re,
  Alina Beygelzimer, Florence d'Alch{\'e} Buc, Emily Fox, and Hugo Larochelle.
\newblock Improving reproducibility in machine learning research (a report from
  the {NeurIPS} 2019 reproducibility program).
\newblock \emph{Journal of Machine Learning Research}, 22\penalty0
  (164):\penalty0 1--20, 2021.

\bibitem[Reed et~al.(2023)Reed, Gupta, Li, Brockman, Funk, Clipp, Keutzer,
  Candido, Uyttendaele, and Darrell]{reed2023scalemae}
Colorado~J. Reed, Ritwik Gupta, Shufan Li, Sarah Brockman, Christopher Funk,
  Brian Clipp, Kurt Keutzer, Salvatore Candido, Matt Uyttendaele, and Trevor
  Darrell.
\newblock {Scale-MAE}: A scale-aware masked autoencoder for multiscale
  geospatial representation learning.
\newblock In \emph{ICCV}, pages 4065--4076. IEEE, 2023.

\bibitem[Robinson(2026)]{robinson2026similar}
Caleb Robinson.
\newblock Similar but different: a {Sentinel-2} benchmark for probing
  spectral-band reliance in land-cover models.
\newblock Hugging Face dataset \texttt{calebrob6/similar-but-different}, 2026.

\bibitem[Rolf et~al.(2021)Rolf, Proctor, Carleton, Bolliger, Shankar, Ishihara,
  Recht, and Hsiang]{rolf2021mosaiks}
Esther Rolf, Jonathan Proctor, Tamma Carleton, Ian Bolliger, Vaishaal Shankar,
  Miyabi Ishihara, Benjamin Recht, and Solomon Hsiang.
\newblock A generalizable and accessible approach to machine learning with
  global satellite imagery.
\newblock \emph{Nature Communications}, 12\penalty0 (1):\penalty0 4392, 2021.

\bibitem[Rustowicz et~al.(2019)Rustowicz, Cheong, Wang, Ermon, Burke, and
  Lobell]{rustowicz2019croptype}
Rose~M. Rustowicz, Robin Cheong, Lijing Wang, Stefano Ermon, Marshall Burke,
  and David~B. Lobell.
\newblock Semantic segmentation of crop type in {Africa}: A novel dataset and
  analysis of deep learning methods.
\newblock In \emph{CVPRW}, pages 75--82, 2019.

\bibitem[Sainte Fare~Garnot and Landrieu(2021)]{garnot2021pastis}
Vivien Sainte Fare~Garnot and Loic Landrieu.
\newblock Panoptic segmentation of satellite image time series with
  convolutional temporal attention networks.
\newblock In \emph{ICCV}, pages 4852--4861. IEEE, 2021.

\bibitem[Sainte Fare~Garnot et~al.(2022)Sainte Fare~Garnot, Landrieu, and
  Chehata]{garnot2022pastisr}
Vivien Sainte Fare~Garnot, Loic Landrieu, and Nesrine Chehata.
\newblock Multi-modal temporal attention models for crop mapping from satellite
  time series.
\newblock \emph{{ISPRS} Journal of Photogrammetry and Remote Sensing},
  187:\penalty0 294--305, 2022.

\bibitem[Si et~al.(2026)Si, Wan, Wang, Do, and Zhao]{si2026chronoearth}
Haozhe Si, Yuxuan Wan, Yuqing Wang, Minh Do, and Han Zhao.
\newblock {ChronoEarth-492K}: A large scale and long horizon spatiotemporal
  hyperspectral earth observation dataset and benchmark, 2026.

\bibitem[Sim{\'e}oni et~al.(2026)Sim{\'e}oni, Vo, Seitzer, Baldassarre, Oquab,
  Jose, Khalidov, Szafraniec, Yi, Ramamonjisoa, Massa, Haziza, Wehrstedt, Wang,
  Darcet, Moutakanni, Sentana, Roberts, Vedaldi, Tolan, Brandt, Couprie,
  Mairal, J{\'e}gou, Labatut, and Bojanowski]{simeoni2025dinov3}
Oriane Sim{\'e}oni, Huy~V. Vo, Maximilian Seitzer, Federico Baldassarre, Maxime
  Oquab, Cijo Jose, Vasil Khalidov, Marc Szafraniec, Seungeun Yi, Micha{\"e}l
  Ramamonjisoa, Francisco Massa, Daniel Haziza, Luca Wehrstedt, Jianyuan Wang,
  Timoth{\'e}e Darcet, Th{\'e}o Moutakanni, Leonel Sentana, Claire Roberts,
  Andrea Vedaldi, Jamie Tolan, John Brandt, Camille Couprie, Julien Mairal,
  Herv{\'e} J{\'e}gou, Patrick Labatut, and Piotr Bojanowski.
\newblock {DINOv3}.
\newblock \emph{Transactions on Machine Learning Research}, 2026.

\bibitem[Simumba et~al.(2026)Simumba, Lehmann, Fraccaro, Alemohammad, De~Mel,
  Khan, Maskey, Longepe, Zhu, Kerner, Bernabe-Moreno, and
  Lacoste]{simumba2025geobench2}
Naomi Simumba, Nils Lehmann, Paolo Fraccaro, Hamed Alemohammad, Geeth De~Mel,
  Salman Khan, Manil Maskey, Nicolas Longepe, Xiao~Xiang Zhu, Hannah Kerner,
  Juan Bernabe-Moreno, and Alexandre Lacoste.
\newblock {GEO-Bench-2}: From performance to capability, rethinking evaluation
  in geospatial {AI}.
\newblock \emph{Transactions on Machine Learning Research}, 2026.

\bibitem[Szwarcman et~al.(2026)Szwarcman, Roy, Fraccaro, G{\'i}slason,
  Blumenstiel, Ghosal, de~Oliveira, de~Sousa~Almeida, Sedona, Kang,
  Chakraborty, Wang, Gomes, Kumar, Gaur, Truong, Godwin, Khallaghi, Lee, Hsu,
  Asanjan, Mujeci, Shidham, Balogun, Kolluru, Keenan, Arevalo, Li, Alemohammad,
  Olofsson, Mayer, Hain, Kennedy, Zadrozny, Bell, Cavallaro, Watson, Maskey,
  Ramachandran, and Moreno]{szwarcman2024prithvi}
Daniela Szwarcman, Sujit Roy, Paolo Fraccaro, {\TH}orsteinn~El{\'i}
  G{\'i}slason, Benedikt Blumenstiel, Rinki Ghosal, Pedro~Henrique de Oliveira,
  Joao~Lucas de Sousa~Almeida, Rocco Sedona, Yanghui Kang, Srija Chakraborty,
  Sizhe Wang, Carlos Gomes, Ankur Kumar, Vishal Gaur, Myscon Truong, Denys
  Godwin, Sam Khallaghi, Hyunho Lee, Chia-Yu Hsu, Ata~Akbari Asanjan, Besart
  Mujeci, Disha Shidham, Rufai~Omowunmi Balogun, Venkatesh Kolluru, Trevor
  Keenan, Paulo Arevalo, Wenwen Li, Hamed Alemohammad, Pontus Olofsson, Timothy
  Mayer, Christopher Hain, Robert Kennedy, Bianca Zadrozny, David Bell,
  Gabriele Cavallaro, Campbell Watson, Manil Maskey, Rahul Ramachandran, and
  Juan~Bernabe Moreno.
\newblock {Prithvi-EO-2.0}: A versatile multitemporal foundation model for
  {Earth} observation applications.
\newblock \emph{IEEE Transactions on Geoscience and Remote Sensing},
  64:\penalty0 1--20, 2026.

\bibitem[Tseng et~al.(2023)Tseng, Cartuyvels, Zvonkov, Purohit, Rolnick, and
  Kerner]{tseng2023presto}
Gabriel Tseng, Ruben Cartuyvels, Ivan Zvonkov, Mirali Purohit, David Rolnick,
  and Hannah Kerner.
\newblock Lightweight, pre-trained transformers for remote sensing timeseries,
  2023.

\bibitem[Tseng et~al.(2025)Tseng, Fuller, Reil, Herzog, Beukema, Bastani,
  Green, Shelhamer, Kerner, and Rolnick]{tseng2025galileo}
Gabriel Tseng, Anthony Fuller, Marlena Reil, Henry Herzog, Patrick Beukema,
  Favyen Bastani, James~R. Green, Evan Shelhamer, Hannah Kerner, and David
  Rolnick.
\newblock Galileo: Learning global \& local features of many remote sensing
  modalities.
\newblock In \emph{ICML}, pages 60280--60300. PMLR, 2025.

\bibitem[Tseng et~al.(2026)Tseng, Bastani, Johnson, Zhang, Wolters, Shah,
  Sablon, Redmon, DeHaven, Ratner, Levy, Steward, Midgley, Howe, Pitelka, and
  Schmitt]{tseng2026olmoearthv13}
Gabriel Tseng, Favyen Bastani, Patrick~Alan Johnson, Yawen Zhang, Piper
  Wolters, Ando Shah, Hadrien Sablon, Joseph Redmon, Heather DeHaven, Rachel
  Ratner, Sasha Levy, Jordan Steward, Katrina Midgley, Andrew Howe, Hunter
  Pitelka, and Ted Schmitt.
\newblock {OlmoEarth} v1.3: Embeddings, 2026.

\bibitem[Wang et~al.(2025)Wang, Xiong, Liu, Stewart, Dujardin, Bountos, Zavras,
  Gerken, Papoutsis, Leal-Taix{\'e}, and Zhu]{wang2025copernicusbench}
Yi Wang, Zhitong Xiong, Chenying Liu, Adam~J. Stewart, Thomas Dujardin,
  Nikolaos~Ioannis Bountos, Angelos Zavras, Franziska Gerken, Ioannis
  Papoutsis, Laura Leal-Taix{\'e}, and Xiao~Xiang Zhu.
\newblock Towards a unified {Copernicus} foundation model for earth vision.
\newblock In \emph{ICCV}, pages 9888--9899. IEEE, 2025.

\bibitem[Xiong et~al.(2024)Xiong, Wang, Zhang, Stewart, Hanna, Borth,
  Papoutsis, Le~Saux, Camps-Valls, and Zhu]{xiong2024dofa}
Zhitong Xiong, Yi Wang, Fahong Zhang, Adam~J. Stewart, Jo{\"e}lle Hanna, Damian
  Borth, Ioannis Papoutsis, Bertrand Le~Saux, Gustau Camps-Valls, and
  Xiao~Xiang Zhu.
\newblock Neural plasticity-inspired multimodal foundation model for earth
  observation, 2024.

\bibitem[Yeh et~al.(2021)Yeh, Meng, Wang, Driscoll, Rozi, Liu, Lee, Burke,
  Lobell, and Ermon]{yeh2021sustainbench}
Christopher Yeh, Chenlin Meng, Sherrie Wang, Anne Driscoll, Erik Rozi, Patrick
  Liu, Jihyeon Lee, Marshall Burke, David~B. Lobell, and Stefano Ermon.
\newblock {SustainBench}: Benchmarks for monitoring the sustainable development
  goals with machine learning.
\newblock In \emph{{NeurIPS} 2021 Track on Datasets and Benchmarks}, 2021.

\bibitem[Young et~al.(2026)Young, Gabtraupov, Soror, and
  Keshav]{young2026usable}
Robin Young, Artyom Gabtraupov, Kenzy Soror, and Srinivasan Keshav.
\newblock How usable are geospatial foundation models? {A} systematic
  evaluation of 89 models, 2026.

\end{thebibliography}

\clearpage
\appendix
\section*{Supplementary Material}
\addcontentsline{toc}{section}{Supplementary Material}
\renewcommand{\theHfigure}{S\arabic{figure}}
\renewcommand{\theHtable}{S\arabic{table}}
\renewcommand{\theHequation}{S\arabic{equation}}
\renewcommand{\thefigure}{S\arabic{figure}}
\renewcommand{\thetable}{S\arabic{table}}
\setcounter{figure}{0}
\setcounter{table}{0}
\renewcommand{\theequation}{S\arabic{equation}}

This supplementary material follows the order of the main paper. Supp.~\ref{sup:benchmark} supports Secs.~\ref{sec:related} and~\ref{sec:benchmark} with an audit of the three benchmark suites closest to ours, the details of the six datasets and the data release. Supp.~\ref{sup:protocol} supports Secs.~\ref{sec:protocol} and~\ref{sec:repro} with the settings of every model, head and metric, the validation of the harness against published figures and the three unstated evaluation choices. Supp.~\ref{sup:results} supports Sec.~\ref{sec:results} with the complete results at every label ratio, the global ranking of every configuration, the inference cost and an analysis of the embedding spaces.

\section{The \paper benchmark}
\label{sup:benchmark}

\subsection{Audit of existing benchmark suites}
\label{sup:suites}
Tables~\ref{tab:supp-geobench}--\ref{tab:supp-pangaea} audit the three suites closest to \paper, dataset by dataset, against its two entry criteria (Sec.~\ref{sec:benchmark}). GEO-Bench~\cite{lacoste2023geobench} has no time axis in its schema, so none of its twelve tasks carries a time series. GEO-Bench-2~\cite{simumba2025geobench2} reads four datasets as multi-temporal, but only a fixed subset of their dates (7 of about 43 on \pastis and 7 of the 12 months of \biomass), each encoded independently, so no model receives a series. PANGAEA~\cite{marsocci2024pangaea} keeps the full series of three datasets, the only ones that meet both criteria, although its readers pass a subset of the dates to the model (six on \pastis and \ctm). We adopt these three from their original releases and read the full year.

This complements a recent audit of the literature~\cite{corley2026noone}, which surveys the benchmarks reported across 152 geospatial foundation model papers and finds 401 distinct ones: a typical paper shares only 27\,\% of its benchmarks with the ten most common, and 35\,\% of papers share none at all. None of those ten carries a year-long \Sone and \Stwo series. Corley~\etal study whether papers converge on the same benchmarks. Our tables go one step further and examine what a model actually receives once a suite is adopted.

\begin{table*}[htbp]
  \centering
  \caption{Audit of GEO-Bench against the two criteria of Sec.~\ref{sec:benchmark}: every Sentinel-2 band and a real time series. \pass{Green}: both criteria hold. \fail{Red}: at least one fails. The suite's schema has no time axis, so every task is single-date.}
  \label{tab:supp-geobench}
  \scriptsize
  \setlength{\tabcolsep}{5pt}
  \begin{tabular}{@{}lllll@{}}
    \toprule
    Dataset & Task & Modality & Temporal axis & Usable \\
    \midrule
    \fail{m-bigearthnet} & Multi-label cls. & Sentinel-2, 12 bands & None & \fail{No} \\
    \fail{m-so2sat} & Classification & Sentinel-2 and Sentinel-1 & None & \fail{No} \\
    \fail{m-brick-kiln} & Classification & Sentinel-2, 13 bands & None & \fail{No} \\
    \fail{m-forestnet} & Classification & Landsat-8, 6 bands & None & \fail{No} \\
    \fail{m-eurosat} & Classification & Sentinel-2, 13 bands & None & \fail{No} \\
    \fail{m-pv4ger} & Classification & Aerial, visible color & None & \fail{No} \\
    \fail{m-pv4ger-seg} & Segmentation & Aerial, visible color & None & \fail{No} \\
    \fail{m-chesapeake-landcover} & Segmentation & Aerial, visible and near infrared & None & \fail{No} \\
    \fail{m-cashew-plantation} & Segmentation & Sentinel-2, 13 bands & None & \fail{No} \\
    \fail{m-SA-crop-type} & Segmentation & Sentinel-2, 13 bands & None & \fail{No} \\
    \fail{m-nz-cattle} & Segmentation & Aerial, visible color & None & \fail{No} \\
    \fail{m-NeonTree} & Segmentation & Visible, hyperspectral and elevation & None & \fail{No} \\
    \bottomrule
  \end{tabular}
  \\[2pt]\raggedright\scriptsize 0 of 12 pass both criteria.
\end{table*}

\begin{table*}[htbp]
  \centering
  \caption{Audit of GEO-Bench-2 against the same two criteria. Four of its nineteen datasets are multi-temporal, but the protocol encodes each date independently, so no model receives a series. The three datasets that carry a full series are already in our benchmark, taken from their original releases.}
  \label{tab:supp-geobench2}
  \scriptsize
  \setlength{\tabcolsep}{5pt}
  \begin{tabular}{@{}lllll@{}}
    \toprule
    Dataset & Task & Modality & Temporal axis & Usable \\
    \midrule
    \fail{BigEarthNet v2} & Multi-label cls. & Sentinel-2, 12 bands, and Sentinel-1 & None & \fail{No} \\
    \fail{TreeSatAI-TS} & Multi-label cls. & Sentinel-2, 10 bands, and Sentinel-1 & Series published & \fail{No}, the protocol sets no dates for it \\
    \fail{So2Sat} & Classification & Sentinel-2 and Sentinel-1 & None & \fail{No} \\
    \fail{ForestNet} & Classification & Landsat-8, 6 bands & None & \fail{No} \\
    \fail{BioMassters} & Regression & Sentinel-2 and Sentinel-1 & 7 of 12 months read & \fail{No}, protocol subsamples \\
    \fail{CaFFe} & Segmentation & SAR, one channel & None & \fail{No} \\
    \fail{CloudSEN12+} & Segmentation & Sentinel-2, 13 bands, and Sentinel-1 & None & \fail{No} \\
    \fail{NASA Burn Scars} & Segmentation & Harmonized Landsat and Sentinel-2, 6 bands & None & \fail{No} \\
    \fail{DynamicEarthNet} & Segmentation & Planet, 4 bands & Daily, 6 dates read & \fail{No}, protocol subsamples \\
    \fail{FLAIR 2} & Segmentation & Aerial, visible and near infrared, elevation & Not read & \fail{No}, its Sentinel-2 series is not read \\
    \fail{Fields of the World} & Segmentation & Sentinel-2, 4 bands & Two dates & \fail{No} \\
    \fail{KuroSiwo} & Segmentation & Sentinel-1 with elevation and slope & Two dates & \fail{No} \\
    \fail{PASTIS-R} & Segmentation & Sentinel-2 and Sentinel-1 & 7 of about 43 dates read & \fail{No}, protocol subsamples \\
    \fail{SpaceNet 2} & Segmentation & WorldView, 8 bands & None & \fail{No} \\
    \fail{SpaceNet 7} & Segmentation & Planet, visible color & Collapsed to one date & \fail{No} \\
    \fail{EverWatch} & Object detection & Aerial, visible color & None & \fail{No} \\
    \fail{m-nz-cattle} & Object detection & Aerial, visible color & None & \fail{No} \\
    \fail{PASTIS-R panoptic} & Instance seg. & Sentinel-2 and Sentinel-1 & 7 of about 43 dates read & \fail{No}, protocol subsamples \\
    \fail{Substations} & Instance seg. & Sentinel-2, 13 bands & Revisits collapsed to one composite & \fail{No} \\
    \bottomrule
  \end{tabular}
  \\[2pt]\raggedright\scriptsize 0 of 19 pass both criteria.
\end{table*}

\begin{table*}[htbp]
  \centering
  \caption{Audit of PANGAEA against the same two criteria. The suite treats temporality as an explicit axis, but its readers subsample the series. Three of the eleven datasets pass both criteria and are the ones we adopted.}
  \label{tab:supp-pangaea}
  \scriptsize
  \setlength{\tabcolsep}{5pt}
  \begin{tabular}{@{}lllll@{}}
    \toprule
    Dataset & Task & Modality & Temporal axis & Usable \\
    \midrule
    \fail{HLS Burn Scars} & Segmentation & Harmonized Landsat and Sentinel-2, 6 bands & None & \fail{No} \\
    \fail{MADOS} & Segmentation & Sentinel-2, 11 bands & None & \fail{No} \\
    \pass{PASTIS-R} & Segmentation & Sentinel-2 and Sentinel-1 & Full series & \pass{Yes} \\
    \fail{Sen1Floods11} & Segmentation & Sentinel-2 and Sentinel-1 & None & \fail{No} \\
    \fail{xView2} & Change detection & Commercial high resolution, visible color & Two dates & \fail{No} \\
    \fail{Five Billion Pixels} & Segmentation & Gaofen-2, 4 bands & None & \fail{No} \\
    \fail{DynamicEarthNet} & Segmentation & Planet, 4 bands & Series over six days & \fail{No}, six days is not a season \\
    \pass{Crop Type Mapping, South Sudan} & Segmentation & Sentinel-2, Sentinel-1 and Planet & Full series & \pass{Yes} \\
    \fail{SpaceNet 7} & Change detection & Planet, visible color & Two dates & \fail{No} \\
    \fail{AI4SmallFarms} & Segmentation & Sentinel-2, 4 bands & None & \fail{No} \\
    \pass{BioMassters} & Regression & Sentinel-2 and Sentinel-1 & Full series & \pass{Yes} \\
    \bottomrule
  \end{tabular}
  \\[2pt]\raggedright\scriptsize 3 of 11 pass both criteria.
\end{table*}

\subsection{Dataset details}
\label{sup:datasets}
\begin{figure*}[t]
  \centering
  \includegraphics[width=\textwidth]{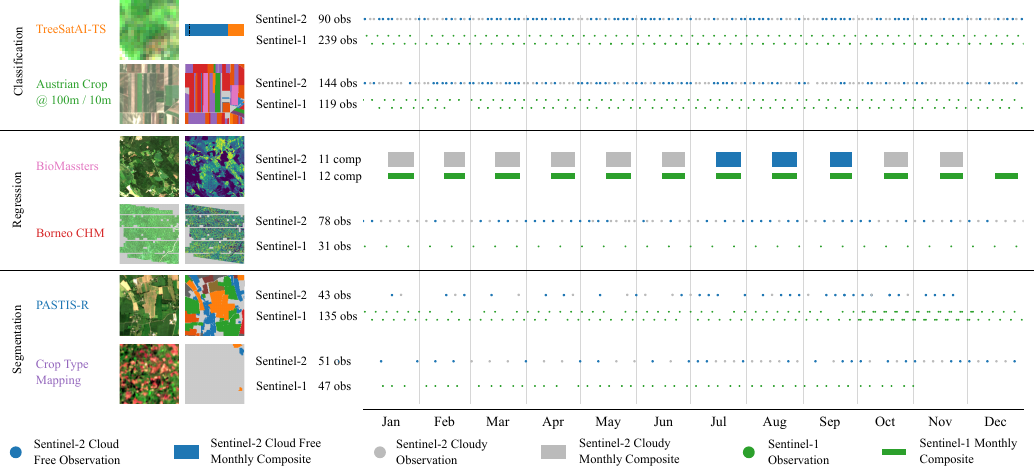}
  \caption{\textbf{Acquisition calendars of the six evaluated datasets.} Each row pairs a representative sample and its annotation with the sample's observation timeline over one year. Observation counts are those of the drawn sample, not the dataset mean, and vary with orbit geometry and acquisition year. \pastis spans two calendar years (September 2018 to November 2019), stacked on the shared twelve-month axis. In \austria, the 10\,m series has 144 dates, of which the 100\,m release keeps the 94 under its 90\,\% cloud threshold.}
  \label{fig:datasets}
\end{figure*}

Fig.~\ref{fig:datasets} shows one sample of each dataset with its annotation and acquisition calendar. Five datasets carry Level-2A surface reflectance, acquired from the Planetary Computer (\austria, \borneo) or as published by their original authors (\treesat, \pastis, \biomass). \ctm uses Level-1C top-of-atmosphere reflectance because its 2017 imagery was acquired before Level-2A products were available in Earth Engine. 

\paragraph{\treesat} A sample is a 6$\times$6 pixel footprint at 10\,m, covering a 60\,m forest plot. The task is multi-label classification of tree genera (e.g.\ \emph{Quercus}, \emph{Picea}, \emph{Fagus}). Each plot lists between one and six genera. A genus counts as present from 7\,\% of the plot, the dataset's own threshold. Predictions are binarized at a probability of 0.5, the published protocol, which is not tuned.

\paragraph{\austria} The 100~m release keeps one pixel in every hundred of the 10~m tile rather than averaging the block. A median parcel covers two samples and most parcels fewer than four. Only a pixel-wise encoder can consume this decimated product as published. A spatial encoder would need the scene re-encoded at 10~m, which is why the 100~m task penalizes every encoder that mixes neighboring pixels. Pixels of the zero class, which marks unlabeled ground, are excluded throughout, and the release names its classes by crop code. For the classification split, fields are assigned to training within each crop code from the smallest upward until their cumulative area reaches the label ratio (Supp.~\ref{sup:unstated}). The remaining fields are shuffled under a fixed seed. One seventh goes to validation and the rest to test. Because the partition changes with the label ratio, there is no full-label column. We report the eight ratios of the reference~\cite{feng2025tessera}, from 1 to 30\,\%. For segmentation the scene is divided into five vertical strips. Test patches come from the topmost, validation patches from the next, and training patches from the remaining three. Our 10~m labels carry 46\,\% more labeled pixels than the reference's 100~m labels, so the two do not score the same ground truth.

\paragraph{\pastis} Only the semantic channel of the annotation is used (the panoptic instance channels are excluded). Our cloud masks are derived from the scene classification of the Planetary Computer \Stwo archive, with thin cirrus and snow kept valid. Co-registration with the release was verified by correlating a reprojected B04 window with the published reflectance of the same patch and date.

\paragraph{\ctm} The four scored crops are sorghum, maize, rice and groundnut. Thirty-one chips with no clear optical observation in the year are excluded from every split. Labels are sparse. About 2\,\% of a chip's pixels carry any annotation and only 1.6\,\% belong to one of the four scored crops. Sorghum accounts for two thirds of the labeled area. A typical chip contributes a few dozen scored pixels. The authors' split was verified chip by chip against the one used by the reference. Chip locations were jittered by up to 3~km before being removed. Its \Stwo reflectance is top-of-atmosphere, while every encoder normalizes with Level-2A statistics. This domain shift adds to the geographic one. We do not correct it because the original release provides only Level-1C imagery.

\paragraph{\biomass} The release publishes no chip position. Its metadata tables list file names and checksums, and its GeoTIFFs carry no georeference. The series therefore cannot be extended with further imagery, and the models that require coordinates are not scored on this dataset (Tab.~\ref{tab:main}).

\paragraph{\borneo} Canopy height ranges from 0 to 71.1~m. Most patches are gap-free although some of them have gaps covering up to 26\,\%. The split is drawn by region because a patch-level split would leak neighboring canopy structure across the boundary. The reference repository provides only a training script and a download link.

\paragraph{Geometry recovery} Each \borneo label patch is a window of the airborne LiDAR canopy-height raster acquired in February 2020 and published through CEDA. We recovered the position of each patch by matching it against that raster. The 18 patches without a confident match were excluded. The \austria scene was located from a preprocessing script that the release does not reference.

\subsection{Data release}
\label{sup:release}

\paper is distributed as a collection of repositories on the Hugging Face Hub (Fig.~\ref{fig:supp-release}a). Each of the six datasets is a separate repository whose card states the task, sample size, acquisition dates, split, metric and license, together with the references of the original release. The leaderboard reports the Score and Mean rank of Eq.~\eqref{eq:score}, with one tab per task reproducing the columns of Tab.~\ref{tab:main} (Fig.~\ref{fig:supp-release}c). The repositories will be made public soon. Fig.~\ref{fig:supp-release}b shows the file trees of two of them.

\paragraph{Licenses} Each redistributed dataset carries the license listed in Tab.~\ref{tab:datasets}. What we acquired or derived (Tab.~\ref{tab:releases}) is released under CC-BY 4.0. The 100\,m bundle of \austria and the published labels of \borneo carry no license statement. They are therefore fetched directly from their origin instead of redistributed. Embeddings follow the license of the model that produced them, stated per folder.

\paragraph{Submission flow} Results are added to the leaderboard through two routes, both open from the public release. In the first, the authors of a model run the released harness and open a pull request with their results and the run configuration file. We check that file against the released protocol. A run that departs from it is either listed with the departure noted or rejected. In the second route, the authors send the embeddings of their model on the six datasets and we fit the heads. Each row is labeled \emph{user submitted} or \emph{measured by the benchmark authors}. This distinction corresponds to the authorship axis of Sec.~\ref{sec:repro}. Each row also records the harness commit, the version of the reference points and the revisions of the repositories it read.

\begin{figure}[!t]
  \centering
  \includegraphics[width=\linewidth]{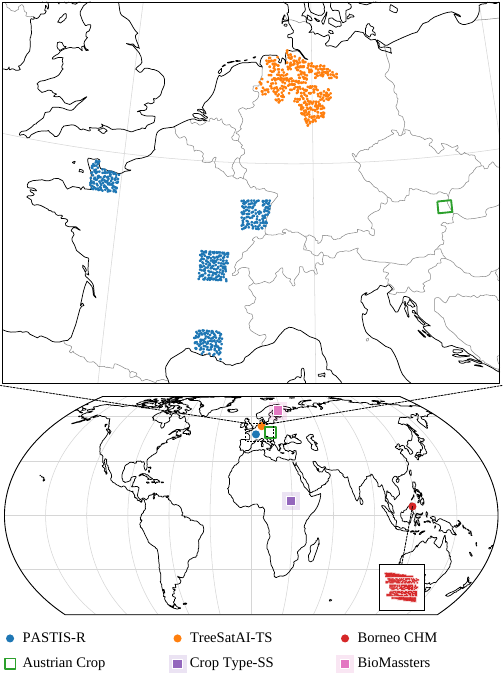}
  \caption{\textbf{Geographic distribution of the \paper benchmark datasets.} Points: per-sample coordinates (\pastis, \treesat, and 482 of 500 recovered \borneo patches; European clusters thinned to one per $0.1^\circ$ cell). Open rectangles: spatial extent (\austria, geo-located from its preprocessing script). Filled squares: regional centroids (\ctm and \biomass, which omit exact coordinates by design). Datasets without coordinates cannot evaluate coordinate-conditioned models, hence the missing columns in Tab.~\ref{tab:main}. Four of the six datasets are European (Sec.~\ref{sec:conclusion}).}
  \label{fig:geometry}
\end{figure}

\begin{figure*}[t]
  \centering
  \includegraphics[width=\textwidth]{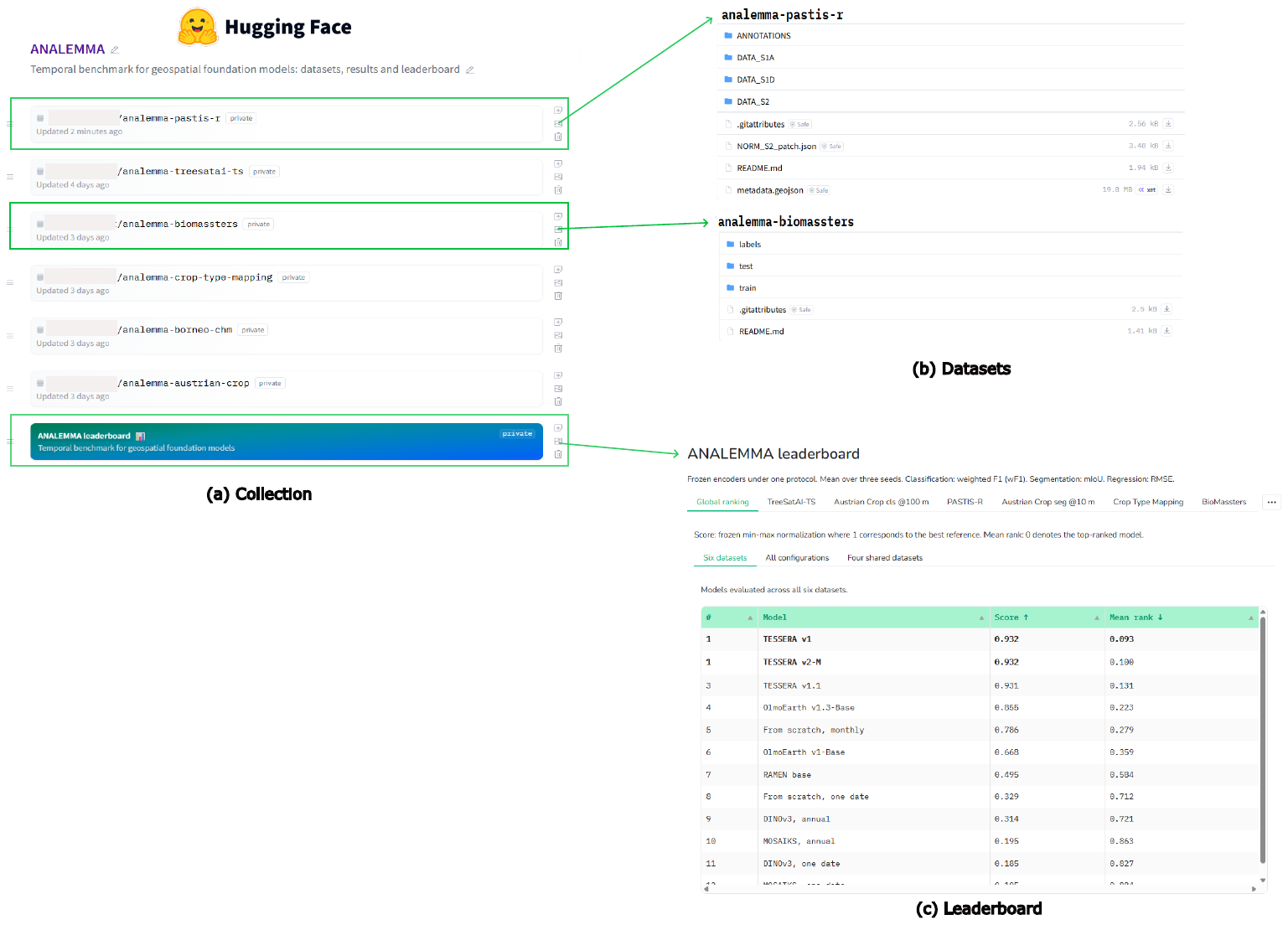}
  \caption{\textbf{Release of \paper on the Hugging Face Hub.} (a)~Dataset collection and leaderboard. (b)~File trees of two repositories. (c)~Global ranking.}
  \label{fig:supp-release}
\end{figure*}

\FloatBarrier
\section{Evaluation protocol}
\label{sup:protocol}

\subsection{Protocol details and metrics}
\label{sup:details}

\begin{figure}[t]
  \centering
  \includegraphics[width=\linewidth]{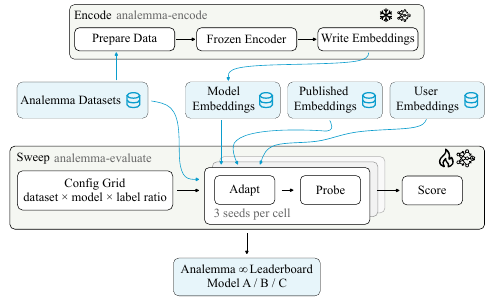}
  \caption{Overview of the \paper evaluation pipeline, from encoding to the global leaderboard. Each cell of the grid (dataset $\times$ model $\times$ label ratio) fits a protocol-defined head on frozen embeddings over three seeds.}
  \label{fig:harness}
\end{figure}

This section completes Sec.~\ref{sec:protocol} with the settings of the encoders, heads, label ratios and metrics. Encoder families follow Tab.~\ref{tab:models}. Each encoder is loaded from its published checkpoint with its released normalization statistics. Monthly composites average the valid observations of each calendar month for every model except \olmothree. A series longer than twelve months, as in \pastis, folds onto one year. Token grids are resampled bilinearly onto the label grid when the head reads them. On \treesat the embedding map is mean-pooled into one vector.

\paragraph{Pixel-wise temporal encoders}
\tessera v1 and v1.1 are two checkpoints of the same model code at different sizes. v2-M is a separate design, the medium student distilled from a 2\,B-parameter teacher~\cite{feng2026tesserav2}. A comparison between v1.1 and v2-M therefore mixes dimensionality with architecture. Their differences cannot be attributed to a single design factor. All three read every acquisition of the ten \Stwo bands at 10 and 20\,m and of both \Sone orbits, one pixel sequence at a time, with a validity mask per observation.

\presto is loaded from the default checkpoint of its repository. Each pixel is a sequence of twelve months with the ten \Stwo bands, their NDVI, the ascending \Sone orbit and the sample coordinates. Months without a valid observation are masked, as are the climate, elevation and land-cover inputs, which our series do not carry.

\alphaearth is read from its annual embedding product on Earth Engine, on the grid of each sample and without further normalization. The year matches the series (2019 on \pastis, 2022 on \austria, 2020 on \borneo and the year of each plot on \treesat). On \austria at 100\,m the 10\,m product is decimated onto the grid of the release, as the release does with its imagery.

\olmothree follows the deployment of Sec.~\ref{sec:protocol} with three differences from the pipeline of its authors. Within a month, observations are ranked by the share of the sample they observe, because our series carry no scene-level cloud cover. The embedding is kept in floating point, without the 8-bit quantization of the released product. Landsat is not provided. Windows without observed tokens are stored as zeros.

\paragraph{Spatio-temporal cube encoders}
\galileo base receives twelve monthly means of thirteen channels (the ten \Stwo bands, NDVI and the two \Sone polarizations), with its location input filled. A patch is tiled into 16~px windows of 4~px tokens, one forward each. These windows are larger than the 12~px windows used during pretraining. A 6~px plot of \treesat is one window of 3~px tokens.

\olmo reads the whole sample in one forward with 4~px tokens (3~px on \treesat and 8~px on \biomass). Its embedding averages the output tokens over the observed months and concatenates the \Stwo and \Sone halves. Unobserved months are masked before attention.

\prithvi 300M reads six bands (B02, B03, B04, B8A, B11 and B12) as four quarterly frames, each the mean of three monthly composites, on the 10\,m grid, although it was pretrained on 30\,m imagery. Its embedding averages the patch tokens of the last block over the four frames. The model takes no temporal mask, so a quarter without a valid observation enters as zero reflectance (0.07\,\% of the frame pixels in a sample of 24 \pastis patches). We report only the plain checkpoint, since the date-conditioned variant performs within the seed spread on every task.

\ramen is loaded from its published checkpoint without the elevation projector and uses the token sizes of \olmo. Its embedding concatenates the \Stwo and \Sone tokens of the last block. \ramen publishes no normalization statistics. Following PANGAEA~\cite{marsocci2024pangaea}, we standardize each dataset with statistics from up to 512 samples of its training split. On \austria at 100\,m we use the whole scene, whose split changes with the label ratio.

\paragraph{Single-image encoders and baselines}
\dino reads the B04, B03 and B02 bands of \Stwo, rendered with a fixed stretch and resized bilinearly to 512~px (224~px on \treesat). A token therefore covers 4~px of a 128~px patch. The \austria scene at 100\,m is a single image of 32$\times$32 tokens. The embedding is the patch token of the last layer. The single date is the observation with the most valid pixels, with ties resolved toward day 196 of the year. The image includes any remaining cloud. The annual variant encodes the mean of all valid observations. The monthly variant encodes each observed month separately and concatenates the twelve embeddings, with unobserved months left as zeros.

\mosaiks applies 2\,048 filters of 4$\times$4~px to the same bands and dates as \dino. The filters are whitened image patches drawn from 64 training patches of \pastis and shared by all datasets. The positive and negative parts of each response are average-pooled over tiles of 8~px (4~px on \borneo and the whole plot on \treesat), where the original method pools over about 1\,km.

\paragraph{Heads and optimization} The classification head is an MLP with hidden layers of 512 and 256 units (batch normalization, ReLU, dropout 0.3). The dense head is a U-Net decoder with 128, 256 and 512 channels (dropout 0.1). All heads are trained with AdamW at a learning rate of $10^{-3}$. The loss is cross-entropy for single-label classification, binary cross-entropy for multi-label targets and mean squared error over valid pixels for regression. We select the epoch with the best validation score. The seed controls the label subsample, the batch order and the head initialization. Pixel classification (\austria at 100\,m) trains for 200 epochs with batches of 8\,192 pixels and weight decay~0.01. Multi-label classification (\treesat) uses batches of 512 plots with the same weight decay and epoch count. Segmentation (\pastis, \austria at 10\,m, \ctm) uses batches of 64 patches, weight decay $10^{-4}$ and a cosine schedule, for 60 epochs on \pastis and \ctm and 100 epochs on \austria at 10\,m. Regression on \borneo uses batches of 16 patches without weight decay for 100 epochs. \biomass is trained for 19\,540 optimizer steps. On \borneo the 1\,\% label ratio keeps three of the 300 training patches, which explains the large spread in that column.

\paragraph{From-scratch model} The model is the head of the frozen protocol trained directly on the bands, with \Stwo reflectance divided by $10^4$. With one date, the input is the ten \Stwo bands of the observation selected for \dino and a validity flag (11 channels). The monthly input stacks twelve monthly means of the ten \Stwo bands and of the two polarizations of each \Sone orbit, each source with its validity flag (17 channels per month, 204 in total). The annual input is a single mean of the same 17 channels. A month without a valid observation is stored as zero and flagged. We add dihedral augmentation, a linear warm-up and gradient clipping, because the default head settings underfit a randomly initialized network.

\paragraph{Label ratios} The default label ratios are 1\,\%, 30\,\% and 100\,\% of the training labels, used on \treesat, \pastis, \biomass and \borneo. Three tasks follow the ratios of their sources. \austria classification uses the eight ratios of the reference, from 1 to 30\,\%, the highest that its split rule admits (Supp.~\ref{sup:datasets}). \austria segmentation uses the three ratios of the reference (1, 30 and 100\,\%) and adds 5 and 10\,\%. At the added ratios, the two split rules of Supp.~\ref{sup:unstated} are compared. \ctm uses 10, 50 and 100\,\%, the ratios that PANGAEA~\cite{marsocci2024pangaea} reports for it.

\paragraph{Metrics} All three metrics are computed on the test split. For the classification metrics, $TP_k$, $FP_k$ and $FN_k$ are the true positives, false positives and false negatives of class $k$. Weighted F1 averages the per-class F1 with the support $n_k$ of each class as its weight:
\begin{equation}
  \mathrm{wF1} = \sum_{k} \frac{n_k}{\sum_j n_j}\,\frac{2\,TP_k}{2\,TP_k+FP_k+FN_k},
  \label{eq:supp-wf1}
\end{equation}
where $n_k$ is the number of labeled test pixels of class $k$ on \austria and the number of test plots where genus $k$ is present on \treesat (Supp.~\ref{sup:datasets}). For mean IoU, void and unlabeled pixels are removed and the counts are accumulated over the whole test split:
\begin{equation}
  \mathrm{mIoU} = \frac{1}{|\mathcal{K}|}\sum_{k\in\mathcal{K}} \frac{TP_k}{TP_k+FP_k+FN_k},
  \label{eq:supp-miou}
\end{equation}
where $\mathcal{K}=\{k : TP_k+FP_k+FN_k>0\}$. RMSE is computed in each group $g$ of test samples over the set $\Omega_g$ of its pixels with a valid label, and averaged over the groups $\mathcal{G}$:
\begin{equation}
  \mathrm{RMSE} = \frac{1}{|\mathcal{G}|}\sum_{g\in\mathcal{G}} \sqrt{\frac{1}{|\Omega_g|}\sum_{i\in\Omega_g}\left(\hat{y}_i-y_i\right)^2}\,,
  \label{eq:supp-rmse}
\end{equation}
where $\hat{y}_i$ and $y_i$ are the predicted and reference values of pixel $i$. A group is one chip on \biomass, the definition of its release, and one batch of 16 test patches on \borneo, the protocol of its reference.

\paragraph{Reporting conventions}
Corley \etal~\cite{corley2026noone} set six expectations for papers that present geospatial foundation models, and \paper \paper meets those that apply to a benchmark. (i)~Weights under a named license do not apply, since \paper releases no model, and every repository states the license of its data and embeddings (Tab.~\ref{tab:datasets}, Supp.~\ref{sup:release}). (ii)~The seven tasks are a shared core evaluation with fixed splits, heads and metrics. (iii)~Every row of Tab.~\ref{tab:main} and of Tabs.~\ref{tab:supp-treesatai-ts}--\ref{tab:supp-borneo-chm} is rerun under our protocol, and figures copied from a publication appear only in Tab.~\ref{tab:supp-validation}, marked as such. (iv)~Every cell reports the mean and standard deviation over three seeds, and the per-seed results are part of the release. (v)~The harness and the leaderboard fit the same heads on submitted embeddings at a pinned protocol version. (vi)~The from-scratch model and the random-feature floor separate the contribution of pretraining from that of the input. This control is partial. The evaluated encoders differ simultaneously in pretraining data, architecture and objective, as do most models in that audit (94 of the 126 papers with identifiable pretraining data use a configuration that no other paper uses). The benchmark therefore ranks models as their authors release them. It does not attribute a difference to a single design choice. For this reason, Sec.~\ref{sec:results} states the three properties of \tessera as a hypothesis. Those expectations concern the papers that propose models. The six requirements of Sec.~\ref{sec:repro} are their counterpart for the benchmark on which the models are scored.

\subsection{Harness validation}
\label{sup:validation}
Before reporting full results in Sec.~\ref{sup:results}, this subsection verifies that, wherever possible, the evaluation pipeline reproduces published figures when run on the same embeddings. The pipeline (Fig.~\ref{fig:harness}) consists of two stages: encoding each sample, and fitting a head on the stored embeddings. Two source releases publish the embeddings from which their reported figures were computed, \austria at 100~m and \borneo~\cite{feng2025tessera}, covering \tessera, \alphaearth and \presto. These embeddings enter the evaluation sweep exactly as submitted embeddings do, testing the second stage in isolation across two independent code paths: an MLP classifier scored by weighted F1 and a U-Net regressor with a validity mask scored by RMSE. Table~\ref{tab:supp-validation} sets our figures beside the published ones, with each row marked as copied or rerun as Corley \etal~\cite{corley2026noone} recommend. It is the only results table in the paper or in this supplement that holds figures copied from a publication. Every row of every other results table is rerun under our protocol, and the published figures quoted in Supp.~\ref{sup:unstated} are identified as such in the text. A star flags the three reruns that lie more than 2 points from the copied figure, which the paragraphs below discuss.

On \austria the published \alphaearth and \presto arrays are one column wider than the label grid (459$\times$519 against 459$\times$518 pixels), while the \tessera array matches it exactly. The source release resamples the wider arrays bilinearly onto the label grid. We apply the same alignment, as it is the only setting under which our figures can be compared with the published ones.

With that alignment, \tessera and \alphaearth track their published curves at all eight label ratios, with largest deviations of 0.88 and 1.09 weighted F1 points. At 30\,\% of the labels they land at 82.10 against 82.09 and at 56.07 against 56.36. \presto agrees within 1.5 points up to 20\,\%. Beyond that threshold the curves separate. The published curve gains 8.15 points between 20 and 30\,\%, the largest single step in any of the three published curves, against 3.58 in ours. The same split, head and metric reproduce the other two curves at that ratio, so the difference appears specific to that published data point rather than to the evaluation pipeline. The release does not contain sufficient information to identify its origin.

On \borneo with all labels, the three models land within 1.1\,m RMSE of their published figures and in the published order. With fewer labels our error is lower than the published one for every model, by up to 1.52\,m for \tessera at 30\,\% and 3.47\,m for \alphaearth at 1\,\%. At 1\,\% only three training patches remain and the seed spread reaches 1.7\,m RMSE, so \tessera and \alphaearth swap places. The harness therefore reproduces the full-label figures and their ordering. The low-label figures depend on a head configuration that the release does not specify (Supp.~\ref{sup:unstated}).

\begin{table*}[t]
  \centering
  \caption{Harness validation on the embeddings published with the source releases, with every row marked as copied or rerun following Corley \etal~\cite{corley2026noone}. $\square$~copied: figure reported by the source~\cite{feng2025tessera} under its own protocol. $\circlearrowright$~rerun: our split, head and metric on the same published embeddings, mean $\pm$ std over three seeds, at the reference's bilinear alignment on \austria. max~$|\Delta|$: largest absolute difference over the label ratios shown. $\star$: rerun more than 2 points from the copied figure.}
  \label{tab:supp-validation}
  \footnotesize
  \setlength{\tabcolsep}{3.5pt}
    \begin{tabular}{@{}llccccccccc@{}}
    \multicolumn{11}{@{}l}{(a) \austria at 100~m, weighted F1 ($\uparrow$)} \\
    \toprule
    Model & Source & 1\,\% & 3\,\% & 5\,\% & 7\,\% & 10\,\% & 15\,\% & 20\,\% & 30\,\% & max $|\Delta|$ \\
    \midrule
    \multirow{2}{*}{\tessera} & $\square$ copied & 66.15 & 71.62 & 74.32 & 77.22 & 78.69 & 80.12 & 80.79 & 82.09 & \multirow{2}{*}{0.88} \\
     & $\circlearrowright$ rerun & 65.77$\pm$0.31 & 71.85$\pm$0.06 & 74.56$\pm$0.16 & 76.34$\pm$0.24 & 77.89$\pm$0.21 & 79.63$\pm$0.11 & 80.31$\pm$0.20 & 82.10$\pm$0.10 & \\
    \addlinespace
    \multirow{2}{*}{\alphaearth} & $\square$ copied & 37.22 & 40.06 & 42.73 & 45.08 & 48.01 & 50.47 & 52.47 & 56.36 & \multirow{2}{*}{1.09} \\
     & $\circlearrowright$ rerun & 37.19$\pm$0.61 & 40.04$\pm$0.57 & 43.82$\pm$0.84 & 45.91$\pm$0.26 & 48.33$\pm$0.26 & 50.42$\pm$0.15 & 52.53$\pm$0.33 & 56.07$\pm$0.28 & \\
    \addlinespace
    \multirow{2}{*}{\presto} & $\square$ copied & 32.74 & 37.63 & 40.26 & 41.50 & 43.94 & 46.01 & 49.74 & 57.89 & \multirow{2}{*}{5.98} \\
     & $\circlearrowright$ rerun & 33.23$\pm$0.79 & 37.21$\pm$0.69 & 39.30$\pm$1.02 & 41.35$\pm$0.97 & 43.71$\pm$0.44 & 45.96$\pm$0.53 & 48.33$\pm$0.33 & 51.91$\pm$0.48$^{\star}$ & \\
    \bottomrule
  \end{tabular}
  \\[6pt]
  \begin{tabular}{@{}llcccc@{}}
    \multicolumn{6}{@{}l}{(b) \borneo, RMSE in m ($\downarrow$)} \\
    \toprule
    Model & Source & 1\,\% & 30\,\% & All & max $|\Delta|$ \\
    \midrule
    \multirow{2}{*}{\tessera} & $\square$ copied & 34.82 & 15.18 & 12.21 & \multirow{2}{*}{1.52} \\
     & $\circlearrowright$ rerun & 33.92$\pm$1.02 & 13.66$\pm$0.20 & 12.72$\pm$0.09 & \\
    \addlinespace
    \multirow{2}{*}{\alphaearth} & $\square$ copied & 36.03 & 20.62 & 16.11 & \multirow{2}{*}{3.47} \\
     & $\circlearrowright$ rerun & 32.56$\pm$1.40$^{\star}$ & 18.22$\pm$0.30$^{\star}$ & 17.20$\pm$0.07 & \\
    \addlinespace
    \multirow{2}{*}{\presto} & $\square$ copied & 36.43 & 20.95 & 17.88 & \multirow{2}{*}{0.66} \\
     & $\circlearrowright$ rerun & 35.99$\pm$1.66 & 20.36$\pm$0.21 & 18.54$\pm$0.72 & \\
    \bottomrule
  \end{tabular}

\end{table*}

\subsection{Unstated evaluation choices}
\label{sup:unstated}
Table~\ref{tab:unstated} summarizes the three evaluation choices identified in Sec.~\ref{sec:repro}. Each was varied in isolation, with encoder, data, head and seeds held fixed, and for the metric variant on the same set of predictions.

\paragraph{Split rule} The \austria supplementary material~\cite{feng2025tessera} describes two partitions in consecutive paragraphs. Classification uses parcels as the splitting unit: within each crop class, fields are assigned to training from the smallest upward until their cumulative area reaches the label ratio. Segmentation divides the scene into five vertical strips, with test patches from the topmost strip, validation patches from the next, and training patches from the remaining three. The released evaluation code covers classification only and implements the parcel rule~\cite{feng2025tessera}. We implement both rules.

Applied to segmentation patches, the parcel rule places every patch in training, validation and test simultaneously, under three label masks, and concentrates the largest parcels in the test set. At 30\,\% of the labels it raises all ten models measured under both rules, by 7.2 (\prithvi) to 23.5 points (\presto), against a top-two gap of 2.8. At 1\,\% the strip rule lands within 0.25 points of the published \tessera and \presto figures (28.11 against 28.20, and 14.50 against 14.25), while the parcel rule falls 5.6 and 11.3 points away. We adopt the strip rule, which reproduces the published figures and, because its partition precedes the label subsample, also yields the full-label column.

\paragraph{Definition of mIoU} The pooled variant gathers all test pixels into one confusion matrix, then averages IoU over classes. The per-sample variant computes mIoU on each patch separately and averages the patch scores: a rare class confined to one patch and missed there scores zero for that patch. Neither the \tessera paper nor its supplementary material states which definition its \pastis figures use, and its released evaluation code contains classification and regression scripts but no segmentation script or IoU computation~\cite{feng2025tessera}. The reference model of \pastis, U-TAE, uses the pooled variant~\cite{garnot2021pastis}.

We record both variants from the same predictions in every \pastis cell. The difference between them is not a constant offset. Pooled minus per-sample, the sign of Tab.~\ref{tab:unstated}, is negative at 1\,\% of the labels for all twelve models ($-$6.2 to $-$2.4 points) and runs from $-$2.5 for \mosaiks to $+$15.6 for \tessera v2-M with all labels, so no single correction factor converts one definition into the other.

\paragraph{Head settings} The source evaluation describes its segmentation head in general terms: a U-Net trained on the embeddings with a standard optimizer and loss~\cite{feng2025tessera}. No training script is released for the \pastis task, so the hyperparameters, class weighting and any standardization of the embeddings are unknown. We compare our default head with a variant that adds two common settings: inverse-square-root class weighting in the loss and per-channel standardization of the embeddings. At 1\,\% of the labels the variant changes the ten models measured under both configurations by $-$0.2 (\prithvi) to $+$12.4 points (\tessera v2-M), against a top-two gap of 0.6. With all labels it lowers eight of the ten by 1.0 to 2.1 points and raises \dino by 0.2 and \mosaiks by 2.1. The sign of the effect depends on the label ratio, and no single configuration favors every model at every ratio. We report the default head for all models, because the published configuration is unknown. At 1\,\% of the labels this is the less favorable choice for eight of the ten models.

\paragraph{Cross-paper disagreements}
Corley \etal~\cite{corley2026noone} find 46 cases in which papers evaluating the same model on the same benchmark disagree by at least 10 points. They name undocumented changes of split, normalization or adaptation recipe as plausible causes, without measuring any of them. The three cases above measure such causes in isolation. With everything else held fixed, the split rule moves a model by 7.2 to 23.5 points, the definition of mIoU by up to 15.6, and the head settings by up to 12.4. Unstated protocol choices can therefore account for disagreements as large as those the audit documents, though we do not claim they explain any specific case. The gaps are structural, a consequence of how evaluations are published rather than of any single release, and the remedy is a protocol released as code (Sec.~\ref{sec:repro}).

\begin{table*}[t]
  \centering
  \caption{Sensitivity of benchmark scores to three unstated protocol choices, each varied in isolation. $\Delta$~\tessera~v1: effect on \tessera v1. $\Delta$~range: range across all models measured under both settings ($n$ in parentheses). Top-2 gap: gap between first and second under this paper's setting (strip split, pooled mIoU, default head).}
  \label{tab:unstated}
  \small
  \setlength{\tabcolsep}{5pt}
    \begin{tabular}{@{}>{\raggedright\arraybackslash}m{3.3cm} >{\raggedright\arraybackslash}m{2.2cm} >{\raggedright\arraybackslash}m{4.3cm} c c c@{}}
    \toprule
    Protocol variant & Evaluation setting & Published specification & $\Delta$ \tessera~v1 & $\Delta$ range ($n$ models) & Top-2 gap \\
    \midrule
    Split rule: parcel vs.\ strip partition & \austria 10\,m, 30\,\% & Both rules described~\cite{feng2025tessera}; released code implements parcel only & 18.6 & 7.2--23.5 (10) & 2.8 \\
    mIoU aggregation: pooled vs.\ per-sample & \pastis, all labels & Not stated~\cite{feng2025tessera}; U-TAE pools~\cite{garnot2021pastis} & 14.8 & $-$2.5--15.6 (12) & 1.5 \\
    Head configuration: weighting, standardization & \pastis, 1\,\% & Not specified~\cite{feng2025tessera} & 7.6 & $-$0.2--12.4 (10) & 0.6 \\
    \bottomrule
  \end{tabular}

\end{table*}

\FloatBarrier
\section{\paper complete results}
\label{sup:results}

\subsection{Per-dataset results}
\label{sup:per-dataset}
Tab.~\ref{tab:main} of the main paper reports each task at two label ratios, the most restrictive and the most permissive. Tables~\ref{tab:supp-treesatai-ts}--\ref{tab:supp-borneo-chm}, placed at the end of this supplement, give the complete results at every label ratio as mean $\pm$ standard deviation over three seeds: \treesat (Tab.~\ref{tab:supp-treesatai-ts}), \austria classification at 100~m (Tab.~\ref{tab:supp-austrian-crop}), \pastis (Tab.~\ref{tab:supp-pastis-r}), \austria segmentation at 10~m (Tab.~\ref{tab:supp-austrian-crop-10m-seg}), \ctm (Tab.~\ref{tab:supp-croptype-ss}), \biomass (Tab.~\ref{tab:supp-biomassters}), and \borneo (Tab.~\ref{tab:supp-borneo-chm}). Rows follow the model families of Tab.~\ref{tab:main}.

\subsection{Complete global ranking}
\label{sup:ranking}
Sec.~\ref{sec:results} normalizes each cell score by min-max scaling (Eq.~\ref{eq:score}) and aggregates the normalized scores hierarchically. The cell-level formula is
\begin{equation}
  s_{m,(t,r)} \;=\; \frac{v_m - v^{\mathrm{w}}}{v^{\mathrm{b}} - v^{\mathrm{w}}}\,,
  \label{eq:supp-score}
\end{equation}
where $v_m$ is the seed mean of model~$m$ in cell $(t,r)$, and $v^{\mathrm{b}}$, $v^{\mathrm{w}}$ are the best and worst seed mean over all configurations in that cell (swapped for RMSE, so that higher is always better). Scores are not clipped to $[0,\,1]$. At present every configuration falls within that range, and a future model that surpasses the current best scores above~1 with no change to the references. The reference points equal, to within 0.01, the best and worst column means of Tabs.~\ref{tab:supp-treesatai-ts}--\ref{tab:supp-borneo-chm}.

\begin{figure}[!htb]
  \centering
  \includegraphics[width=\linewidth]{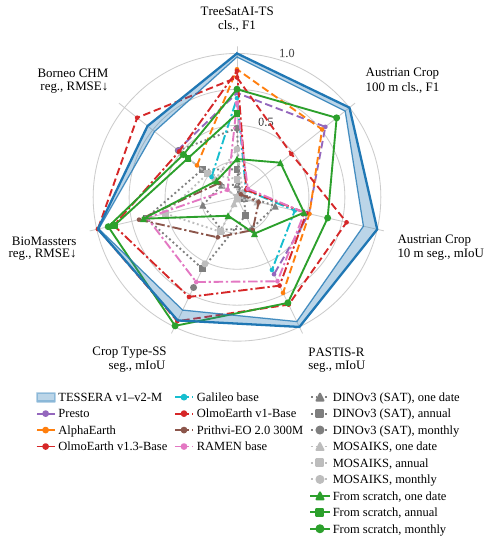}
  \caption{Per-task normalized scores of every configuration ranked in Figs.~\ref{fig:supp-rank-full} and~\ref{fig:supp-rank-shared}, plotted against frozen reference points (1~=~reference best, 0~=~reference worst). A broken line indicates a missing entry on the adjacent axis. \mosaiks lies near~0 on several axes because its configurations set the reference worst.}
  \label{fig:supp-radar}
\end{figure}

\begin{figure*}[t]
  \centering
  \includegraphics[width=\textwidth]{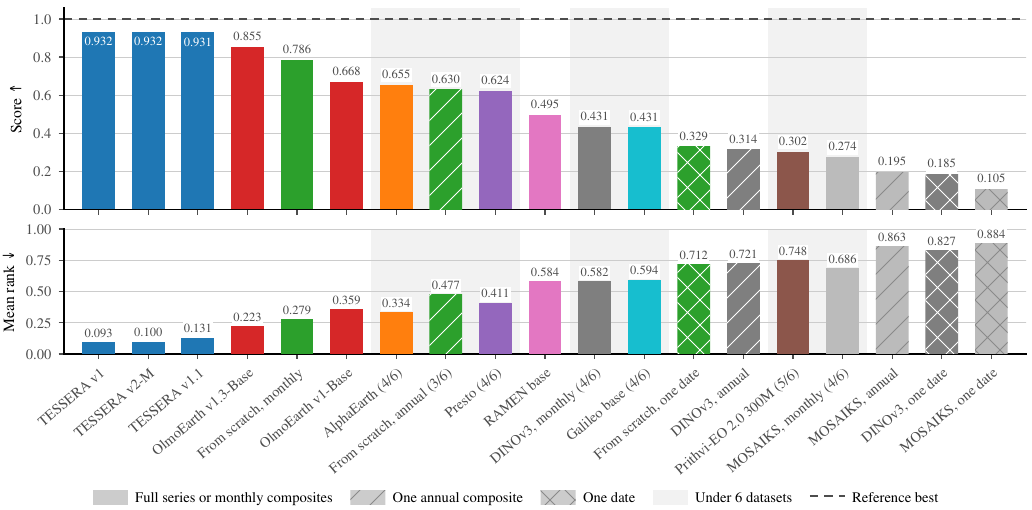}  %
  \caption{Global ranking of every evaluated configuration on the seven tasks of Tab.~\ref{tab:main}. Parentheses and shading mark configurations covering fewer than six datasets. Score and mean rank follow Fig.~\ref{fig:rank}; different support is not directly comparable.}
  \label{fig:supp-rank-full}
\end{figure*}

\begin{figure*}[t]
  \centering
  \includegraphics[width=\textwidth]{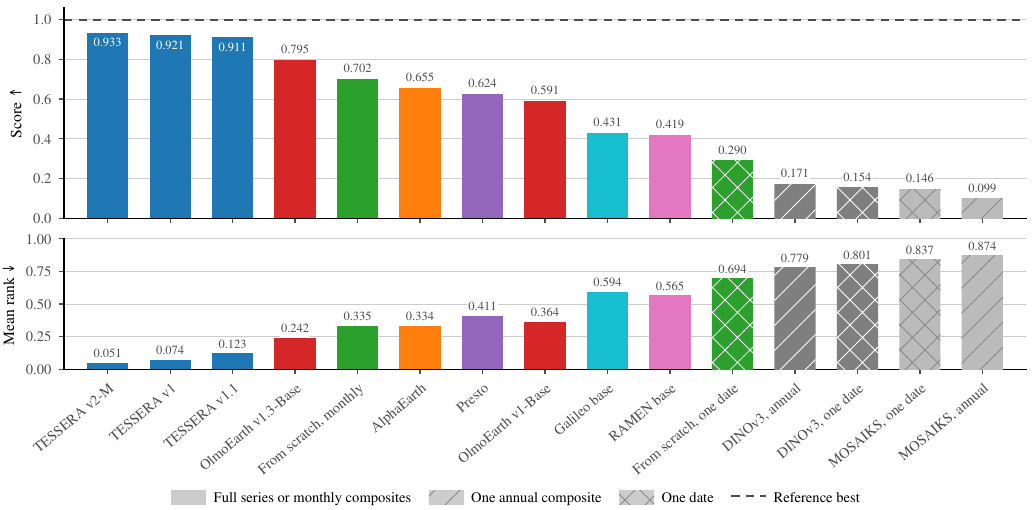}  %
  \caption{Global ranking on the four datasets with per-sample coordinates, on which every encoder can be scored. Score and mean rank follow Fig.~\ref{fig:rank}, restricted to these four datasets.}
  \label{fig:supp-rank-shared}
\end{figure*}

Fig.~\ref{fig:supp-radar} shows the per-task normalized scores underlying the global ranking, plotted against the frozen reference points.  Fig.~\ref{fig:supp-rank-full} aggregates these scores into a global ranking for every evaluated configuration, with the number of datasets each covers given in parentheses when fewer than six; rows with different support are not directly comparable.  Fig.~\ref{fig:supp-rank-shared} ranks only the configurations that cover the four datasets with per-sample coordinates (\treesat, \austria, \pastis, and \borneo), so that coordinate-conditioned encoders absent from Fig.~\ref{fig:rank} can be compared on equal terms.

Sec.~\ref{sec:conclusion} reports the Score on a third subset, the four datasets that were not released with \tessera (\treesat, \pastis, \ctm and \biomass). This subset differs from that of Fig.~\ref{fig:supp-rank-shared}, which keeps the four datasets with per-sample coordinates and therefore includes the two released with \tessera. On the four datasets not released with it, the Score is 0.970, 0.964 and 0.962 for \tessera v1.1, v2-M and v1, 0.902 for \olmothree, 0.869 for the from-scratch model on monthly composites and 0.802 for \olmo. On the other two (\austria and \borneo) it is 0.873, 0.867 and 0.854 for \tessera v1, v2-M and v1.1, 0.760 for \olmothree and 0.621 for the from-scratch model. The three \tessera checkpoints, \olmothree and the from-scratch model keep that order on both subsets and on the six datasets together (Fig.~\ref{fig:rank}).

These rankings compare models under one protocol on six datasets, as Corley \etal~\cite{corley2026noone} advocate. That audit argues that a literature-wide state of the art cannot be established from published results, and we do not claim one. The reference points and per-seed results are released, so a new model can be scored against the same cells.


\FloatBarrier
\subsection{Inference cost}
\label{sup:cost}
Sec.~\ref{sec:results} notes that the from-scratch model on monthly composites is the cheapest configuration at its score.  This subsection details the measurement and expands the comparison.

We measure inference cost, not estimate it. For each dataset we run one forward pass of the encoder and fitted head on a fixed reference sample with all observations valid. Operations are counted with \texttt{torch.utils.flop\_counter} (matrix multiplication, convolution and attention, one multiply-add as two FLOPs). The count covers every forward pass the protocol requires, twelve for a monthly configuration and one for an annual one. Pretraining and head fitting are excluded as one-time costs.  The count is normalized to FLOPs per pixel per year of input on each benchmark's 10\,m grid, so that samples of different spatial and temporal extent are comparable on a single axis.

\begin{figure*}[t]
  \centering
  \includegraphics[width=\textwidth]{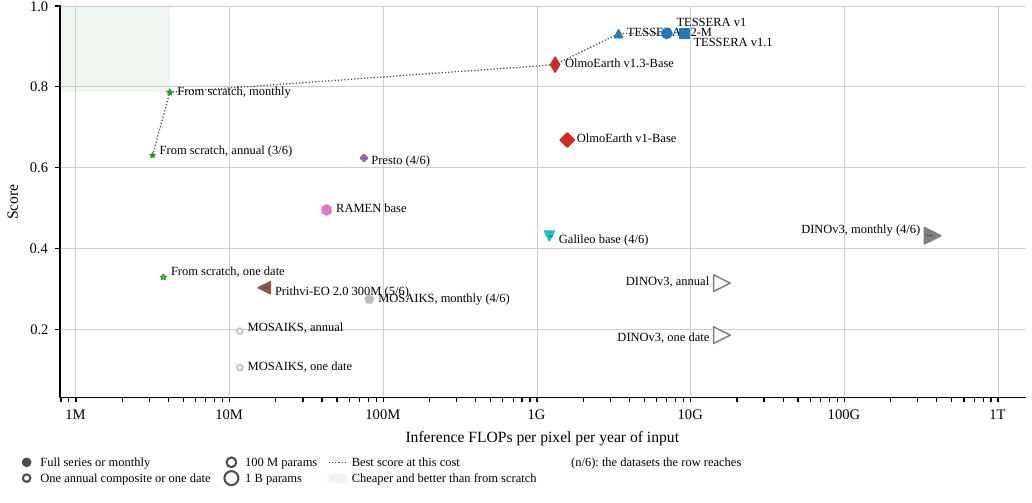}
  \caption{Score vs.\ inference cost (FLOPs per pixel per year of input, log scale). Marker area is proportional to the parameters one inference runs through (encoder plus head, log scale). Filled markers denote the full series or monthly composites, hollow markers one annual composite or one date. The dotted line is the Pareto frontier. The shaded region, cheaper and better than the from-scratch monthly model, is empty. Configurations covering fewer than six datasets show their count in parentheses. \alphaearth is omitted (unpublished encoder).}
  \label{fig:supp-cost}
\end{figure*}

Fig.~\ref{fig:supp-cost} plots the resulting cost against the global score (Tab.~\ref{tab:supp-cost}).  At a mean cost of 4.08\,M FLOPs, the from-scratch model on monthly composites is the cheapest configuration at its score level: no evaluated model scores higher at a lower cost.  The four configurations that do score higher, the three \tessera generations and \olmothree at 1.31\,G mean FLOPs, cost two to three orders of magnitude more.  The shaded region of the figure, cheaper and better than the from-scratch monthly model, is empty.

\begin{table*}[t]
  \centering
  \caption{Inference cost per dataset (FLOPs per pixel per year of input). Mean is the arithmetic mean over the datasets a configuration covers, plotted in Fig.~\ref{fig:supp-cost}. Params: encoder plus head, averaged over datasets. ---: the configuration does not cover the dataset. \alphaearth is omitted (unpublished encoder). Counting methodology is described in the text.}
  \label{tab:supp-cost}
  \footnotesize
  \setlength{\tabcolsep}{4pt}
    \begin{tabular}{@{}llrrrrrrrr@{}}
    \toprule
    Model & Input & \treesat & \austria & \pastis & \ctm & \biomass & \borneo & Mean & Params \\
    \midrule
    \tessera v1 & full year & 7.10\,G & 7.09\,G & 5.92\,G & 7.17\,G & 7.72\,G & 7.07\,G & 7.01\,G & 114\,M \\
    \tessera v1.1 & full year & 18.5\,G & 14.3\,G & 8.29\,G & 5.60\,G & 1.85\,G & 5.97\,G & 9.09\,G & 84\,M \\
    \tessera v2-M & full year & 6.97\,G & 5.34\,G & 3.09\,G & 2.07\,G & 685\,M & 2.21\,G & 3.40\,G & 47\,M \\
    \presto & 12 months & 74.8\,M & 79.3\,M & 66.2\,M & --- & --- & 79.1\,M & 74.9\,M & 24\,M \\
    \olmothree & 12 months & 1.15\,G & 1.45\,G & 1.21\,G & 1.47\,G & 1.14\,G & 1.45\,G & 1.31\,G & 154\,M \\
    \midrule
    \galileo base & 12 months & 1.72\,G & 1.16\,G & 971\,M & --- & --- & 967\,M & 1.20\,G & 110\,M \\
    \olmo & 12 months & 707\,M & 3.46\,G & 2.89\,G & 1.17\,G & 636\,M & 583\,M & 1.57\,G & 116\,M \\
    \prithvi 300M & 4 quarters & --- & 16.8\,M & 14.0\,M & 16.8\,M & 19.6\,M & 16.9\,M & 16.8\,M & 336\,M \\
    \ramen & 12 months & 63.2\,M & 49.3\,M & 41.2\,M & 42.7\,M & 19.5\,M & 40.5\,M & 42.7\,M & 123\,M \\
    \midrule
    \dino & one date & 76.2\,G & 903\,M & 754\,M & 3.61\,G & 257\,M & 14.2\,G & 16.0\,G & 6.7\,B \\
    \dino & annual & 76.2\,G & 903\,M & 754\,M & 3.61\,G & 257\,M & 14.2\,G & 16.0\,G & 6.7\,B \\
    \dino & monthly & 914\,G & --- & --- & 43.3\,G & --- & 170\,G & 376\,G & 6.8\,B \\
    \mosaiks & one date & 174\,k & 14.2\,M & 11.8\,M & 14.3\,M & 15.4\,M & 14.1\,M & 11.7\,M & 30\,M \\
    \mosaiks & annual & 174\,k & 14.2\,M & 11.8\,M & 14.3\,M & 15.4\,M & 14.1\,M & 11.7\,M & 30\,M \\
    \mosaiks & monthly & 2.01\,M & --- & --- & 122\,M & --- & 120\,M & 81.1\,M & 67\,M \\
    \midrule
    From scratch & one date & 7.86\,k & 4.52\,M & 3.77\,M & 4.56\,M & 4.92\,M & 4.50\,M & 3.71\,M & 26\,M \\
    From scratch & annual & 8.03\,k & --- & --- & --- & 4.93\,M & 4.52\,M & 3.15\,M & 21\,M \\
    From scratch & monthly & 13.4\,k & 4.96\,M & 4.14\,M & 5.02\,M & 5.41\,M & 4.95\,M & 4.08\,M & 26\,M \\
    \bottomrule
  \end{tabular}

\end{table*}

\FloatBarrier
\subsection{Embedding analysis}
\label{sup:embedding}
\label{sup:qual}%

\begin{figure*}[t]
  \centering
  \includegraphics[width=\textwidth]{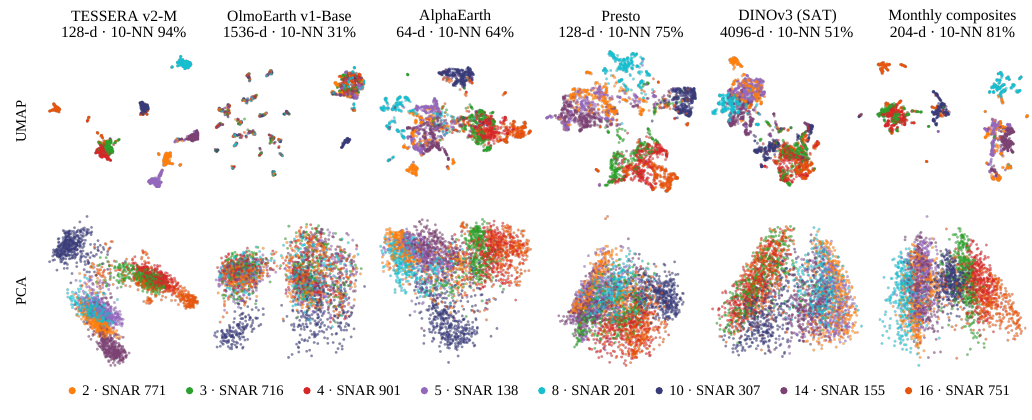}
  \caption{Embedding spaces on \austria at 10~m. Each column is one encoder or the raw monthly composites, labeled with dimensionality and 10-NN accuracy. Each point is a single-class 4$\times$4~px block from the test patches, colored by crop class. Top: UMAP. Bottom: PCA onto the first two components.}
  \label{fig:supp-embedding}
\end{figure*}
Fig.~\ref{fig:supp-embedding} visualizes the embedding space of each encoder on \austria at 10~m, using the same 3\,200 points for every model: 4$\times$4~px blocks of a single class, 400 per class for the eight most frequent classes, sampled from the held-out patches of the test strip. Each block is read at its first pixel for the per-pixel encoders and for the raw composites, and at the covering token for \olmo and \dino, whose 4$\times$4~px token matches the block exactly, so no resampling is needed.

Each embedding is projected to two dimensions with two methods. UMAP~\cite{mcinnes2018umap} (top row) is non-linear and can separate classes that are not linearly separable in the original space, so taken alone it would overstate what an encoder has learned. PCA (bottom row) projects linearly onto the two directions of largest variance and reveals how much of that separation is already linear. Neither projection yields a scalar measure. The quantitative evaluation is the accuracy of a 10-nearest-neighbor (10-NN) classifier in each encoder's full-dimensional space, estimated with five-fold cross-validation grouped by patch so that neighboring pixels from a single parcel never appear on both sides of a fold.

The last column applies the same procedure to the raw monthly composites that the from-scratch model receives, with no encoder. At 81\,\%, their 10-NN accuracy falls below \tessera v2-M (94\,\%) and above the other four encoders shown (\presto 75\,\%, \alphaearth 64\,\%, \dino 51\,\% and \olmo 31\,\%). Among the encoders the figure covers, this is the representation-level counterpart of the finding in Sec.~\ref{sec:results} that the input explains most of the performance. \tessera v1 and v1.1, \galileo, \prithvi, \ramen and \olmothree are not shown, the last two because the figure was drawn before they joined the evaluation grid. The figure therefore makes no statement about \olmothree, the only encoder other than \tessera that ranks above the from-scratch model (Fig.~\ref{fig:rank}).
The UMAP of \olmo fragments into islands that mix crop types and patches but share a position within the 128-pixel window: 12.8\,\% of each point's ten nearest neighbors occupy the same position in a different patch, compared with 0.09\,\% expected by chance and at most 0.1\,\% for every other column. The grid visible in the \olmo panel of Fig.~\ref{fig:qual} is the map-view counterpart: the stored embeddings carry the 1.28~km patch borders, and tokens straddling a border differ roughly 2.6$\times$ more than tokens within one patch.

\clearpage
\begin{table*}[htbp]
  \centering
  \caption{TreeSatAI-TS, multi-label classification. Weighted F1 ($\uparrow$) at every label ratio, mean $\pm$ std over three seeds. \best{Bold}: best mean. \second{Underlined}: second best.}
  \label{tab:supp-treesatai-ts}
  \scriptsize
  \setlength{\tabcolsep}{5pt}
  \renewcommand{\arraystretch}{1.3}
  \begin{tabular}{@{}llccc@{}}
    \toprule
    Model & Input & 1\,\% & 30\,\% & All \\
    \midrule
    TESSERA v1 & full year & \second{59.25 $\pm$ 0.40} & \second{72.49 $\pm$ 0.30} & \second{75.55 $\pm$ 0.16} \\
    \addlinespace[2pt]
    TESSERA v1.1 & full year & 58.91 $\pm$ 0.45 & 72.16 $\pm$ 0.33 & 75.46 $\pm$ 0.20 \\
    \addlinespace[2pt]
    TESSERA v2-M & full year & \best{61.09 $\pm$ 0.24} & \best{73.12 $\pm$ 0.47} & \best{76.00 $\pm$ 0.41} \\
    \addlinespace[2pt]
    AlphaEarth & annual product & 53.73 $\pm$ 0.43 & 67.81 $\pm$ 0.59 & 71.35 $\pm$ 0.26 \\
    \addlinespace[2pt]
    Presto & 12 monthly & 43.59 $\pm$ 0.92 & 59.71 $\pm$ 0.75 & 63.34 $\pm$ 0.32 \\
    \addlinespace[2pt]
    OlmoEarth v1.3-Base & 12 monthly & 49.12 $\pm$ 0.65 & 65.55 $\pm$ 0.02 & 69.44 $\pm$ 0.21 \\
    \midrule
    Galileo & 12 monthly & 41.59 $\pm$ 0.76 & 56.96 $\pm$ 0.59 & 62.46 $\pm$ 0.32 \\
    \addlinespace[2pt]
    OlmoEarth v1-Base & 12 monthly & 52.38 $\pm$ 0.14 & 67.54 $\pm$ 0.44 & 71.36 $\pm$ 0.18 \\
    \addlinespace[2pt]
    RAMEN base & 12 monthly & 40.24 $\pm$ 0.79 & 55.38 $\pm$ 0.15 & 59.68 $\pm$ 0.25 \\
    \midrule
    \multirow{3}{*}{DINOv3} & one date & 20.02 $\pm$ 4.23 & 20.18 $\pm$ 0.27 & 20.04 $\pm$ 0.45 \\
     & monthly & 28.80 $\pm$ 0.17 & 46.50 $\pm$ 0.30 & 52.25 $\pm$ 0.14 \\
     & annual & 21.88 $\pm$ 2.64 & 27.59 $\pm$ 0.35 & 29.35 $\pm$ 0.85 \\
    \addlinespace[2pt]
    \multirow{3}{*}{MOSAIKS} & one date & 15.26 $\pm$ 3.84 & 17.07 $\pm$ 1.33 & 13.60 $\pm$ 0.22 \\
     & monthly & 22.16 $\pm$ 0.30 & 37.77 $\pm$ 0.19 & 43.83 $\pm$ 0.33 \\
     & annual & 21.09 $\pm$ 1.14 & 23.98 $\pm$ 0.96 & 21.74 $\pm$ 0.89 \\
    \midrule
    \multirow{3}{*}{From scratch} & one date & 20.62 $\pm$ 2.43 & 34.19 $\pm$ 0.17 & 37.10 $\pm$ 0.57 \\
     & monthly & 38.70 $\pm$ 0.97 & 64.53 $\pm$ 0.30 & 69.53 $\pm$ 0.25 \\
     & annual & 34.06 $\pm$ 0.17 & 53.58 $\pm$ 0.35 & 56.18 $\pm$ 0.12 \\
    \bottomrule
  \end{tabular}
\end{table*}

\begin{table*}[htbp]
  \centering
  \caption{Austrian Crop at 100 m, classification. Weighted F1 ($\uparrow$) at every label ratio, mean $\pm$ std over three seeds. \best{Bold}: best mean. \second{Underlined}: second best. ``n/a'' denotes a configuration that does not apply.}
  \label{tab:supp-austrian-crop}
  \scriptsize
  \setlength{\tabcolsep}{5pt}
  \renewcommand{\arraystretch}{1.3}
  \begin{tabular}{@{}llcccccccc@{}}
    \toprule
    Model & Input & 1\,\% & 3\,\% & 5\,\% & 7\,\% & 10\,\% & 15\,\% & 20\,\% & 30\,\% \\
    \midrule
    TESSERA v1 & full year & \second{69.03 $\pm$ 0.14} & \second{74.12 $\pm$ 0.42} & \second{76.33 $\pm$ 0.34} & \second{78.05 $\pm$ 0.24} & \second{79.61 $\pm$ 0.05} & \second{80.92 $\pm$ 0.14} & \second{81.75 $\pm$ 0.12} & 83.10 $\pm$ 0.04 \\
    \addlinespace[2pt]
    TESSERA v1.1 & full year & 65.98 $\pm$ 0.12 & 73.29 $\pm$ 0.26 & 75.10 $\pm$ 0.44 & 77.08 $\pm$ 0.12 & 78.88 $\pm$ 0.14 & 80.52 $\pm$ 0.22 & 81.44 $\pm$ 0.18 & \second{83.18 $\pm$ 0.13} \\
    \addlinespace[2pt]
    TESSERA v2-M & full year & \best{71.19 $\pm$ 0.65} & \best{75.63 $\pm$ 0.17} & \best{77.30 $\pm$ 0.10} & \best{79.00 $\pm$ 0.19} & \best{80.41 $\pm$ 0.22} & \best{81.38 $\pm$ 0.07} & \best{82.62 $\pm$ 0.16} & \best{84.19 $\pm$ 0.15} \\
    \addlinespace[2pt]
    AlphaEarth & annual product & 53.03 $\pm$ 0.39 & 59.45 $\pm$ 0.39 & 63.12 $\pm$ 0.22 & 65.23 $\pm$ 0.27 & 67.59 $\pm$ 0.31 & 70.57 $\pm$ 0.19 & 71.98 $\pm$ 0.31 & 74.42 $\pm$ 0.03 \\
    \addlinespace[2pt]
    Presto & 12 monthly & 55.66 $\pm$ 0.29 & 61.25 $\pm$ 0.50 & 64.18 $\pm$ 0.56 & 66.80 $\pm$ 0.08 & 69.56 $\pm$ 0.54 & 72.32 $\pm$ 0.31 & 72.92 $\pm$ 0.51 & 75.34 $\pm$ 0.35 \\
    \addlinespace[2pt]
    OlmoEarth v1.3-Base & 12 monthly & 38.62 $\pm$ 0.40 & 43.79 $\pm$ 0.46 & 47.38 $\pm$ 0.52 & 48.82 $\pm$ 1.07 & 52.46 $\pm$ 0.39 & 54.99 $\pm$ 0.85 & 57.52 $\pm$ 0.38 & 61.29 $\pm$ 0.26 \\
    \midrule
    Galileo & 12 monthly & 23.98 $\pm$ 0.38 & 25.63 $\pm$ 0.23 & 26.70 $\pm$ 0.42 & 27.42 $\pm$ 0.19 & 27.92 $\pm$ 0.33 & 29.69 $\pm$ 0.23 & 30.59 $\pm$ 0.43 & 32.43 $\pm$ 0.11 \\
    \addlinespace[2pt]
    OlmoEarth v1-Base & 12 monthly & 23.20 $\pm$ 0.37 & 25.73 $\pm$ 0.56 & 27.69 $\pm$ 0.92 & 28.06 $\pm$ 0.38 & 29.04 $\pm$ 0.15 & 30.08 $\pm$ 0.45 & 30.74 $\pm$ 0.47 & 31.40 $\pm$ 0.59 \\
    \addlinespace[2pt]
    Prithvi-EO 2.0 & 4 quarterly & 23.50 $\pm$ 0.44 & 24.47 $\pm$ 0.38 & 25.77 $\pm$ 0.08 & 25.88 $\pm$ 0.37 & 25.28 $\pm$ 0.80 & 26.03 $\pm$ 0.28 & 26.10 $\pm$ 0.41 & 26.40 $\pm$ 0.40 \\
    \addlinespace[2pt]
    RAMEN base & 12 monthly & 26.01 $\pm$ 0.05 & 26.01 $\pm$ 0.42 & 27.74 $\pm$ 0.33 & 28.41 $\pm$ 0.29 & 30.25 $\pm$ 0.03 & 30.02 $\pm$ 0.45 & 30.98 $\pm$ 0.10 & 32.03 $\pm$ 0.19 \\
    \midrule
    \multirow{3}{*}{DINOv3} & one date & 24.13 $\pm$ 0.24 & 24.81 $\pm$ 0.06 & 25.48 $\pm$ 0.31 & 25.63 $\pm$ 0.42 & 25.37 $\pm$ 0.34 & 26.08 $\pm$ 0.17 & 26.08 $\pm$ 0.15 & 26.31 $\pm$ 0.28 \\
     & monthly & 21.81 $\pm$ 0.02 & 24.16 $\pm$ 0.45 & 24.97 $\pm$ 0.52 & 25.03 $\pm$ 0.77 & 25.87 $\pm$ 0.16 & 25.86 $\pm$ 0.25 & 26.42 $\pm$ 0.16 & 26.24 $\pm$ 0.26 \\
     & annual & 23.10 $\pm$ 0.64 & 24.80 $\pm$ 0.27 & 25.28 $\pm$ 0.33 & 25.71 $\pm$ 0.21 & 25.46 $\pm$ 0.17 & 25.64 $\pm$ 0.21 & 25.93 $\pm$ 0.10 & 26.11 $\pm$ 0.54 \\
    \addlinespace[2pt]
    \multirow{3}{*}{MOSAIKS} & one date & 22.64 $\pm$ 0.77 & 23.21 $\pm$ 0.42 & 23.71 $\pm$ 0.57 & 23.97 $\pm$ 0.47 & 23.47 $\pm$ 0.85 & 24.54 $\pm$ 0.77 & 24.61 $\pm$ 0.66 & 24.68 $\pm$ 0.07 \\
     & monthly & 24.25 $\pm$ 0.38 & 24.73 $\pm$ 0.49 & 26.55 $\pm$ 0.39 & 26.66 $\pm$ 0.30 & 26.68 $\pm$ 0.14 & 27.63 $\pm$ 0.32 & 26.91 $\pm$ 0.50 & 27.60 $\pm$ 0.54 \\
     & annual & 22.04 $\pm$ 0.17 & 22.60 $\pm$ 0.04 & 23.59 $\pm$ 0.68 & 23.81 $\pm$ 0.39 & 23.86 $\pm$ 0.62 & 24.96 $\pm$ 0.59 & 24.08 $\pm$ 0.71 & 24.47 $\pm$ 0.97 \\
    \midrule
    \multirow{3}{*}{From scratch} & one date & 40.53 $\pm$ 0.13 & 42.33 $\pm$ 0.26 & 43.67 $\pm$ 0.07 & 45.13 $\pm$ 0.14 & 46.14 $\pm$ 0.36 & 46.62 $\pm$ 0.21 & 47.15 $\pm$ 0.20 & 48.15 $\pm$ 0.51 \\
     & monthly & 60.32 $\pm$ 0.35 & 67.12 $\pm$ 0.04 & 70.43 $\pm$ 0.14 & 72.45 $\pm$ 0.14 & 74.81 $\pm$ 0.07 & 77.58 $\pm$ 0.14 & 78.68 $\pm$ 0.04 & 81.08 $\pm$ 0.13 \\
     & annual & n/a & n/a & n/a & n/a & n/a & n/a & n/a & n/a \\
    \bottomrule
  \end{tabular}
\end{table*}

\begin{table*}[htbp]
  \centering
  \caption{PASTIS-R, segmentation. mIoU ($\uparrow$) at every label ratio, mean $\pm$ std over three seeds. \best{Bold}: best mean. \second{Underlined}: second best. ``---'' denotes a configuration that was not run, ``n/a'' one that does not apply.}
  \label{tab:supp-pastis-r}
  \scriptsize
  \setlength{\tabcolsep}{5pt}
  \renewcommand{\arraystretch}{1.3}
  \begin{tabular}{@{}llccc@{}}
    \toprule
    Model & Input & 1\,\% & 30\,\% & All \\
    \midrule
    TESSERA v1 & full year & 27.88 $\pm$ 1.76 & \second{59.35 $\pm$ 0.27} & \second{65.17 $\pm$ 0.37} \\
    \addlinespace[2pt]
    TESSERA v1.1 & full year & \second{28.27 $\pm$ 1.60} & 59.08 $\pm$ 0.59 & 64.83 $\pm$ 0.18 \\
    \addlinespace[2pt]
    TESSERA v2-M & full year & \best{28.86 $\pm$ 1.63} & \best{61.58 $\pm$ 0.26} & \best{66.67 $\pm$ 0.06} \\
    \addlinespace[2pt]
    AlphaEarth & annual product & 22.82 $\pm$ 2.19 & 49.06 $\pm$ 0.36 & 54.96 $\pm$ 0.16 \\
    \addlinespace[2pt]
    Presto & 12 monthly & 18.21 $\pm$ 2.48 & 42.32 $\pm$ 0.51 & 51.42 $\pm$ 0.18 \\
    \addlinespace[2pt]
    OlmoEarth v1.3-Base & 12 monthly & 24.73 $\pm$ 3.05 & 53.32 $\pm$ 0.67 & 59.47 $\pm$ 0.44 \\
    \midrule
    Galileo & 12 monthly & 16.21 $\pm$ 3.21 & 41.06 $\pm$ 1.37 & 52.55 $\pm$ 0.13 \\
    \addlinespace[2pt]
    OlmoEarth v1-Base & 12 monthly & 19.28 $\pm$ 2.63 & 46.95 $\pm$ 1.14 & 57.24 $\pm$ 0.35 \\
    \addlinespace[2pt]
    Prithvi-EO 2.0 & 4 quarterly & 12.29 $\pm$ 1.28 & 26.23 $\pm$ 0.41 & 31.28 $\pm$ 0.26 \\
    \addlinespace[2pt]
    RAMEN base & 12 monthly & 20.00 $\pm$ 1.96 & 44.71 $\pm$ 0.70 & 52.78 $\pm$ 0.07 \\
    \midrule
    \multirow{3}{*}{DINOv3} & one date & 11.97 $\pm$ 1.01 & 25.04 $\pm$ 0.68 & 31.40 $\pm$ 0.19 \\
     & monthly & --- & --- & --- \\
     & annual & 9.51 $\pm$ 0.90 & 21.49 $\pm$ 0.36 & 25.97 $\pm$ 0.21 \\
    \addlinespace[2pt]
    \multirow{3}{*}{MOSAIKS} & one date & 8.07 $\pm$ 0.77 & 14.19 $\pm$ 0.34 & 18.11 $\pm$ 0.23 \\
     & monthly & --- & --- & --- \\
     & annual & 8.30 $\pm$ 1.24 & 13.19 $\pm$ 0.81 & 17.31 $\pm$ 0.12 \\
    \midrule
    \multirow{3}{*}{From scratch} & one date & 9.05 $\pm$ 0.35 & 30.27 $\pm$ 0.99 & 39.04 $\pm$ 0.20 \\
     & monthly & 22.14 $\pm$ 1.31 & 54.34 $\pm$ 0.53 & 62.60 $\pm$ 0.42 \\
     & annual & n/a & n/a & n/a \\
    \bottomrule
  \end{tabular}
\end{table*}

\begin{table*}[htbp]
  \centering
  \caption{Austrian Crop at 10 m, segmentation. mIoU ($\uparrow$) at every label ratio, mean $\pm$ std over three seeds. \best{Bold}: best mean. \second{Underlined}: second best. ``---'' denotes a configuration that was not run, ``n/a'' one that does not apply.}
  \label{tab:supp-austrian-crop-10m-seg}
  \scriptsize
  \setlength{\tabcolsep}{5pt}
  \renewcommand{\arraystretch}{1.3}
  \begin{tabular}{@{}llccccc@{}}
    \toprule
    Model & Input & 1\,\% & 5\,\% & 10\,\% & 30\,\% & All \\
    \midrule
    TESSERA v1 & full year & 28.11 $\pm$ 0.55 & \second{35.40 $\pm$ 1.53} & \second{46.01 $\pm$ 1.26} & \second{50.61 $\pm$ 1.90} & \second{58.19 $\pm$ 1.99} \\
    \addlinespace[2pt]
    TESSERA v1.1 & full year & \second{28.14 $\pm$ 1.48} & 35.02 $\pm$ 2.00 & 45.22 $\pm$ 1.55 & 49.41 $\pm$ 3.33 & 56.01 $\pm$ 0.28 \\
    \addlinespace[2pt]
    TESSERA v2-M & full year & \best{31.38 $\pm$ 0.62} & \best{40.17 $\pm$ 0.56} & \best{48.97 $\pm$ 2.51} & \best{53.43 $\pm$ 2.79} & \best{59.88 $\pm$ 2.14} \\
    \addlinespace[2pt]
    AlphaEarth & annual product & 15.33 $\pm$ 1.90 & 22.22 $\pm$ 1.03 & 28.82 $\pm$ 0.68 & 30.93 $\pm$ 0.19 & 35.44 $\pm$ 1.27 \\
    \addlinespace[2pt]
    Presto & 12 monthly & 14.50 $\pm$ 0.54 & 21.20 $\pm$ 1.27 & 28.85 $\pm$ 1.26 & 31.23 $\pm$ 0.55 & 37.18 $\pm$ 0.88 \\
    \addlinespace[2pt]
    OlmoEarth v1.3-Base & 12 monthly & 23.66 $\pm$ 0.50 & 31.21 $\pm$ 1.61 & 41.11 $\pm$ 1.56 & 43.50 $\pm$ 0.42 & 49.15 $\pm$ 1.50 \\
    \midrule
    Galileo & 12 monthly & 8.14 $\pm$ 0.61 & 16.68 $\pm$ 1.18 & 24.89 $\pm$ 0.47 & 28.86 $\pm$ 0.44 & 36.35 $\pm$ 0.10 \\
    \addlinespace[2pt]
    OlmoEarth v1-Base & 12 monthly & 10.96 $\pm$ 1.32 & 19.81 $\pm$ 0.72 & 28.41 $\pm$ 1.21 & 32.24 $\pm$ 1.29 & 40.61 $\pm$ 0.22 \\
    \addlinespace[2pt]
    Prithvi-EO 2.0 & 4 quarterly & 6.39 $\pm$ 0.58 & 8.98 $\pm$ 0.25 & 10.67 $\pm$ 0.63 & 12.53 $\pm$ 0.71 & 16.19 $\pm$ 0.20 \\
    \addlinespace[2pt]
    RAMEN base & 12 monthly & 12.17 $\pm$ 0.95 & 19.96 $\pm$ 1.82 & 27.15 $\pm$ 0.19 & 31.07 $\pm$ 1.06 & 37.09 $\pm$ 0.66 \\
    \midrule
    \multirow{3}{*}{DINOv3} & one date & 9.83 $\pm$ 1.06 & 13.41 $\pm$ 0.38 & 16.81 $\pm$ 1.10 & 18.80 $\pm$ 0.63 & 20.77 $\pm$ 0.38 \\
     & monthly & --- & --- & --- & --- & --- \\
     & annual & 3.79 $\pm$ 0.37 & 5.47 $\pm$ 0.25 & 6.63 $\pm$ 0.46 & 7.47 $\pm$ 0.29 & 8.52 $\pm$ 0.05 \\
    \addlinespace[2pt]
    \multirow{3}{*}{MOSAIKS} & one date & 6.80 $\pm$ 0.54 & 8.71 $\pm$ 0.85 & 10.95 $\pm$ 0.30 & 11.17 $\pm$ 0.65 & 13.65 $\pm$ 0.11 \\
     & monthly & --- & --- & --- & --- & --- \\
     & annual & 3.32 $\pm$ 0.55 & 3.81 $\pm$ 0.04 & 4.37 $\pm$ 0.16 & 4.03 $\pm$ 0.63 & 4.78 $\pm$ 0.65 \\
    \midrule
    \multirow{3}{*}{From scratch} & one date & 10.47 $\pm$ 1.62 & 20.70 $\pm$ 1.29 & 27.78 $\pm$ 1.19 & 31.47 $\pm$ 0.77 & 37.50 $\pm$ 1.42 \\
     & monthly & 16.39 $\pm$ 0.32 & 25.84 $\pm$ 2.45 & 34.24 $\pm$ 0.95 & 39.93 $\pm$ 2.41 & 46.86 $\pm$ 1.17 \\
     & annual & n/a & n/a & n/a & n/a & n/a \\
    \bottomrule
  \end{tabular}
\end{table*}

\begin{table*}[htbp]
  \centering
  \caption{Crop Type Mapping, South Sudan, segmentation. mIoU ($\uparrow$) at every label ratio, mean $\pm$ std over three seeds. \best{Bold}: best mean. \second{Underlined}: second best. ``n/a'' denotes a configuration that does not apply.}
  \label{tab:supp-croptype-ss}
  \scriptsize
  \setlength{\tabcolsep}{5pt}
  \renewcommand{\arraystretch}{1.3}
  \begin{tabular}{@{}llccc@{}}
    \toprule
    Model & Input & 10\,\% & 50\,\% & All \\
    \midrule
    TESSERA v1 & full year & 36.73 $\pm$ 2.15 & \second{59.83 $\pm$ 5.53} & 59.72 $\pm$ 10.75 \\
    \addlinespace[2pt]
    TESSERA v1.1 & full year & 40.52 $\pm$ 4.78 & 58.19 $\pm$ 2.16 & 59.69 $\pm$ 0.51 \\
    \addlinespace[2pt]
    TESSERA v2-M & full year & 37.43 $\pm$ 11.04 & 51.87 $\pm$ 2.88 & 60.85 $\pm$ 4.05 \\
    \addlinespace[2pt]
    OlmoEarth v1.3-Base & 12 monthly & \best{42.00 $\pm$ 7.84} & 57.70 $\pm$ 2.18 & 58.73 $\pm$ 3.93 \\
    \midrule
    OlmoEarth v1-Base & 12 monthly & 30.47 $\pm$ 9.01 & 49.98 $\pm$ 0.92 & \second{60.86 $\pm$ 1.25} \\
    \addlinespace[2pt]
    Prithvi-EO 2.0 & 4 quarterly & 18.15 $\pm$ 1.77 & 31.82 $\pm$ 4.70 & 43.80 $\pm$ 4.68 \\
    \addlinespace[2pt]
    RAMEN base & 12 monthly & 25.02 $\pm$ 5.38 & 46.94 $\pm$ 1.71 & 58.40 $\pm$ 2.84 \\
    \midrule
    \multirow{3}{*}{DINOv3} & one date & 14.99 $\pm$ 1.27 & 34.04 $\pm$ 3.28 & 39.68 $\pm$ 2.76 \\
     & monthly & 27.40 $\pm$ 1.64 & 47.01 $\pm$ 5.81 & 59.76 $\pm$ 1.97 \\
     & annual & 33.13 $\pm$ 1.06 & 39.03 $\pm$ 8.45 & 43.59 $\pm$ 8.16 \\
    \addlinespace[2pt]
    \multirow{3}{*}{MOSAIKS} & one date & 18.91 $\pm$ 1.67 & 19.90 $\pm$ 0.76 & 25.21 $\pm$ 1.50 \\
     & monthly & 24.77 $\pm$ 4.46 & 40.03 $\pm$ 4.46 & 49.73 $\pm$ 1.70 \\
     & annual & 19.68 $\pm$ 4.98 & 29.87 $\pm$ 5.20 & 38.83 $\pm$ 5.37 \\
    \midrule
    \multirow{3}{*}{From scratch} & one date & 19.85 $\pm$ 0.46 & 26.40 $\pm$ 9.25 & 28.29 $\pm$ 3.66 \\
     & monthly & \second{41.33 $\pm$ 6.82} & \best{59.99 $\pm$ 7.82} & \best{61.49 $\pm$ 10.71} \\
     & annual & n/a & n/a & n/a \\
    \bottomrule
  \end{tabular}
\end{table*}

\begin{table*}[htbp]
  \centering
  \caption{BioMassters, regression. RMSE ($\downarrow$) at every label ratio, mean $\pm$ std over three seeds. \best{Bold}: best mean. \second{Underlined}: second best. ``---'' denotes a configuration that was not run.}
  \label{tab:supp-biomassters}
  \scriptsize
  \setlength{\tabcolsep}{5pt}
  \renewcommand{\arraystretch}{1.3}
  \begin{tabular}{@{}llccc@{}}
    \toprule
    Model & Input & 1\,\% & 30\,\% & All \\
    \midrule
    TESSERA v1 & full year & 33.55 $\pm$ 0.17 & \best{30.61 $\pm$ 0.06} & \best{30.53 $\pm$ 0.01} \\
    \addlinespace[2pt]
    TESSERA v1.1 & full year & \second{33.32 $\pm$ 0.03} & \second{30.64 $\pm$ 0.09} & \second{30.60 $\pm$ 0.12} \\
    \addlinespace[2pt]
    TESSERA v2-M & full year & 33.57 $\pm$ 0.01 & 30.75 $\pm$ 0.05 & 30.75 $\pm$ 0.11 \\
    \addlinespace[2pt]
    OlmoEarth v1.3-Base & 12 monthly & \best{32.74 $\pm$ 0.12} & 30.89 $\pm$ 0.10 & 30.86 $\pm$ 0.10 \\
    \midrule
    OlmoEarth v1-Base & 12 monthly & 37.76 $\pm$ 0.20 & 33.04 $\pm$ 0.04 & 32.94 $\pm$ 0.10 \\
    \addlinespace[2pt]
    Prithvi-EO 2.0 & 4 quarterly & 44.58 $\pm$ 0.33 & 37.28 $\pm$ 0.05 & 36.42 $\pm$ 0.14 \\
    \addlinespace[2pt]
    RAMEN base & 12 monthly & 43.39 $\pm$ 0.21 & 39.73 $\pm$ 0.04 & 39.39 $\pm$ 0.12 \\
    \midrule
    \multirow{3}{*}{DINOv3} & one date & 55.93 $\pm$ 0.42 & 49.75 $\pm$ 0.14 & 48.55 $\pm$ 0.07 \\
     & monthly & --- & --- & --- \\
     & annual & 44.56 $\pm$ 0.49 & 39.12 $\pm$ 0.15 & 38.46 $\pm$ 0.05 \\
    \addlinespace[2pt]
    \multirow{3}{*}{MOSAIKS} & one date & 61.81 $\pm$ 0.39 & 55.94 $\pm$ 0.16 & 55.78 $\pm$ 0.31 \\
     & monthly & --- & --- & --- \\
     & annual & 47.64 $\pm$ 0.88 & 42.58 $\pm$ 0.09 & 42.81 $\pm$ 0.63 \\
    \midrule
    \multirow{3}{*}{From scratch} & one date & 41.85 $\pm$ 0.65 & 39.38 $\pm$ 0.09 & 39.01 $\pm$ 0.09 \\
     & monthly & 35.78 $\pm$ 0.02 & 32.61 $\pm$ 0.08 & 32.09 $\pm$ 0.09 \\
     & annual & 36.52 $\pm$ 0.19 & 34.12 $\pm$ 0.04 & 33.33 $\pm$ 0.05 \\
    \bottomrule
  \end{tabular}
\end{table*}

\begin{table*}[htbp]
  \centering
  \caption{Borneo CHM, regression. RMSE ($\downarrow$) at every label ratio, mean $\pm$ std over three seeds. \best{Bold}: best mean. \second{Underlined}: second best.}
  \label{tab:supp-borneo-chm}
  \scriptsize
  \setlength{\tabcolsep}{5pt}
  \renewcommand{\arraystretch}{1.3}
  \begin{tabular}{@{}llccc@{}}
    \toprule
    Model & Input & 1\,\% & 30\,\% & All \\
    \midrule
    TESSERA v1 & full year & 34.11 $\pm$ 1.04 & 14.19 $\pm$ 0.30 & \best{13.06 $\pm$ 0.22} \\
    \addlinespace[2pt]
    TESSERA v1.1 & full year & 34.31 $\pm$ 1.37 & \second{14.13 $\pm$ 0.06} & 13.25 $\pm$ 0.14 \\
    \addlinespace[2pt]
    TESSERA v2-M & full year & 35.19 $\pm$ 0.64 & \best{14.10 $\pm$ 0.18} & \second{13.08 $\pm$ 0.06} \\
    \addlinespace[2pt]
    AlphaEarth & annual product & 34.27 $\pm$ 2.80 & 17.75 $\pm$ 0.51 & 16.63 $\pm$ 0.17 \\
    \addlinespace[2pt]
    Presto & 12 monthly & 35.18 $\pm$ 1.61 & 15.62 $\pm$ 0.31 & 15.01 $\pm$ 0.22 \\
    \addlinespace[2pt]
    OlmoEarth v1.3-Base & 12 monthly & \second{31.44 $\pm$ 0.67} & 14.66 $\pm$ 0.24 & 13.76 $\pm$ 0.03 \\
    \midrule
    Galileo & 12 monthly & 35.22 $\pm$ 0.31 & 18.49 $\pm$ 0.33 & 17.06 $\pm$ 0.11 \\
    \addlinespace[2pt]
    OlmoEarth v1-Base & 12 monthly & 34.31 $\pm$ 2.02 & 16.36 $\pm$ 0.39 & 15.37 $\pm$ 0.17 \\
    \addlinespace[2pt]
    Prithvi-EO 2.0 & 4 quarterly & 35.27 $\pm$ 1.23 & 18.71 $\pm$ 0.05 & 17.92 $\pm$ 0.21 \\
    \addlinespace[2pt]
    RAMEN base & 12 monthly & 35.04 $\pm$ 1.36 & 19.46 $\pm$ 0.16 & 18.64 $\pm$ 0.19 \\
    \midrule
    \multirow{3}{*}{DINOv3} & one date & 34.68 $\pm$ 1.83 & 19.57 $\pm$ 0.59 & 18.02 $\pm$ 0.29 \\
     & monthly & 36.12 $\pm$ 1.38 & 15.28 $\pm$ 0.38 & 14.47 $\pm$ 0.09 \\
     & annual & 35.35 $\pm$ 1.62 & 17.38 $\pm$ 0.40 & 16.68 $\pm$ 0.12 \\
    \addlinespace[2pt]
    \multirow{3}{*}{MOSAIKS} & one date & \best{30.84 $\pm$ 8.64} & 17.84 $\pm$ 0.38 & 17.67 $\pm$ 0.10 \\
     & monthly & 36.11 $\pm$ 0.55 & 17.85 $\pm$ 0.33 & 17.32 $\pm$ 0.04 \\
     & annual & 35.54 $\pm$ 0.42 & 17.59 $\pm$ 0.18 & 17.04 $\pm$ 0.21 \\
    \midrule
    \multirow{3}{*}{From scratch} & one date & 36.31 $\pm$ 0.61 & 18.48 $\pm$ 1.24 & 16.70 $\pm$ 0.04 \\
     & monthly & 35.65 $\pm$ 1.57 & 16.13 $\pm$ 0.27 & 14.85 $\pm$ 0.20 \\
     & annual & 34.44 $\pm$ 2.01 & 17.30 $\pm$ 0.53 & 15.58 $\pm$ 0.40 \\
    \bottomrule
  \end{tabular}
\end{table*}

\end{document}